\documentclass[journal]{IEEEtran}
\usepackage[explicit]{titlesec}
\usepackage{float} 
\usepackage[figuresright]{rotating}
\usepackage{graphicx}
\usepackage{verbatim}
\usepackage{epstopdf}
\usepackage{subcaption}
\usepackage[export]{adjustbox} 
\usepackage{algorithmic}
\usepackage{amsmath}
\usepackage{amsthm}
\usepackage{mathrsfs}
\usepackage{extarrows}
\usepackage{cite}

\usepackage{bbm}
\usepackage{amssymb}
\usepackage{booktabs}

\newcommand{\RNum}[1]{\uppercase\expandafter{\romannumeral #1\relax}}

\usepackage{stfloats}

\usepackage{color, soul}
\usepackage{tikz,xcolor,hyperref}
\hypersetup{colorlinks=true, linkcolor=blue, citecolor=blue, urlcolor=blue,}
\usepackage{multirow}
\usepackage{makecell}

\definecolor{lime}{HTML}{A6CE39}

\makeatletter
\DeclareRobustCommand{\orcidicon}{%
	\begin{tikzpicture}
	\draw[lime, fill=lime] (0,0) 
	circle [radius=0.16] 
	node[white] {{\fontfamily{qag}\selectfont \tiny ID}};    \draw[white, fill=white] (-0.0625,0.095) 
	circle [radius=0.007];    \end{tikzpicture}
	\hspace{-2mm}}
\foreach \x in {A, ..., Z}{%
	\expandafter\xdef\csname orcid\x\endcsname{\noexpand\href{https://orcid.org/\csname orcidauthor\x\endcsname}{\noexpand\orcidicon}}
}

\newcommand*\bigcdot{\mathpalette\bigcdot@{.5}}
\newcommand*\bigcdot@[2]{\mathbin{\vcenter{\hbox{\scalebox{#2}{$\m@th#1\bullet$}}}}}
\makeatother

\usepackage[linesnumbered,ruled]{algorithm2e}

\begin{document}
\title{

Robust Real-Time Coordination of CAVs: \\A Distributed Optimization Framework \\under Uncertainty
    }
\author{Haojie Bai, 
	\emph{}
    Tingting Zhang,
    \emph{}
	 Cong Guo, 
	 \emph{}
	 Yang Wang, 
	 \emph{}
     Xiongwei Zhao\, 
	 \emph{} and Hai Zhu
    \emph{} 
	\vspace{-0.5em}
\thanks{This work was supported in part by the Natural Science Foundation of China under Grant No. 62171160; in part by Science and Technology Project of Shenzhen under Grant JCYJ20200109113424990; and in part by Marine Economy Development Project of Guangdong Province under Grant GDNRC [2020]014. (Corresponding author: Tingting Zhang.)
	
Haojie Bai, Tingting Zhang, Cong Guo, Xiongwei Zhao and Yang Wang are with the School of Electronic and Information Engineering, Harbin Institute of Technology (Shenzhen), Shenzhen 518071, China (e-mail: yangw@hit.edu.cn).

Hai Zhu is with the Defense Innovation Institute, Chinese Academy of Military Sciences, Beijing 100071, China (e-mail: zhuhai11@alumni.nudt.edu.cn).
	
}}

\maketitle
\allowdisplaybreaks
\begin{abstract}

Achieving both safety guarantees and real-time performance in cooperative vehicle coordination remains a fundamental challenge, particularly in dynamic and uncertain environments. Existing methods often suffer from insufficient uncertainty treatment in safety modeling, which intertwines with the heavy computational burden under complex multi-vehicle coupling. This paper presents a novel coordination framework that resolves this challenge through three key innovations: 1) direct control of vehicles' trajectory distributions during coordination, formulated as a robust cooperative planning problem with adaptive enhanced safety constraints, ensuring a specified level of safety regarding the uncertainty of the interactive trajectory, 2) a fully parallel ADMM-based distributed trajectory negotiation (ADMM-DTN) algorithm that efficiently solves the optimization problem while allowing configurable negotiation rounds to balance solution quality and computational resources, and 3) an interactive attention mechanism that selectively focuses on critical interactive participants to further enhance computational efficiency. 
Simulation results demonstrate that our framework achieves significant advantages in safety (reducing collision rates by up to 40.79\% in various scenarios) and real-time performance compared to representative benchmarks, while maintaining strong scalability with increasing vehicle numbers. The proposed interactive attention mechanism further reduces the computational demand by 15.4\%. Real-world experiments further validate robustness and real-time feasibility with unexpected dynamic obstacles, demonstrating reliable coordination in complex traffic scenes.

The experiment demo could be found at https://youtu.be/4PZwBnCsb6Q.

\end{abstract}

\IEEEpeerreviewmaketitle

\section{Introduction}\label{sectionI}

Connected and autonomous vehicles (CAVs) represent a transformative solution to critical transportation challenges, offering the potential to significantly reduce traffic accidents and congestion through coordinated decision-making\cite{eskandarian2019research,sun2021survey}. By integrating advanced sensing, computing, and vehicle-to-everything (V2X) communication capabilities, CAVs can share real-time information and make collaborative decisions that surpass the limitations of individual vehicle operations\cite{zhou2020evolutionary}. This technological convergence enables unprecedented opportunities for system-wide optimization of traffic flow and safety in intelligent transportation systems (ITS)\cite{khayatian2020survey,rios2016survey}.

A fundamental challenge in realizing the full potential of CAVs lies in achieving reliable coordination in critical traffic scenarios, particularly intersections where multiple vehicles must safely negotiate shared space in real time\cite{hult2016coordination}. While existing approaches can generate cooperative trajectories that satisfy nominal safety conditions and traffic rules\cite{wang2019survey,meng2017analysis}, two critical real-world challenges remain unresolved. First, ensuring robust safety under environmental uncertainties remains technically challenging. In practice, vehicles face various sources of uncertainty and interference\cite{knaup2023safe}, including unavoidable disturbances in controllers and actuators, as well as onboard sensor noise that corrupts measurements (e.g., GNSS, radar and inertial measurements)\cite{li2018high,8795887}. These uncertainties can accumulate during vehicle movement, leading to potentially dangerous deviations from planned trajectories and increasing collision risks during vehicle interactions\cite{okamoto2019optimal}. Second, the computational burden of coordination becomes prohibitive as the number of vehicles increases. Real-time trajectory planning for multiple vehicles in spatiotemporal domains requires solving complex optimization problems under tight time constraints, especially challenging with limited onboard computational resources.

These challenges create a fundamental tension between safety and real-time performance in CAV coordination systems. Traditional conservative approaches that prioritize safety often result in overly cautious behavior and reduced traffic efficiency, while faster but less robust methods may compromise safety guarantees. To this end, this paper addresses the tension by developing a novel framework that simultaneously ensures probabilistic safety guarantees and efficient real-time implementation.

\subsection{Related Works}
There are two primary avenues for cooperative trajectory planning design: \emph{learning-based} and \emph{optimization-based} approaches\cite{claussmann2019review}. Learning-based approaches, such as reinforcement learning and imitation learning, can capture complicated driving strategies and continuously refine their proficiency using real-world driving data\cite{guan2020centralized,luo2023real}. Specifically, \cite{chen2023deep,shao2023vehicular} develop an efficient deep reinforcement learning framework for intersection crossing in which CAVs collaboratively learn a driving policy to maximize traffic throughput. However, these data-driven methods lack the theoretical interpretability of neural networks and real-world cooperative driving data, which raises safety concerns. The other approach is optimization-based. This method typically formulates the cooperative planning problem as a mathematical program subject to system constraints arising from vehicle dynamics, collision avoidance, and physical limits, while maximizing certain objectives. Several studies\cite{luo2023computationally, malikopoulos2021optimal,xu2019grouping,bai2023context,pei2019cooperative} obtain optimal or suboptimal coordination solutions using well-known tools and algorithms from optimization theory such as optimal control\cite{malikopoulos2021optimal}, dynamic programming\cite{pei2019cooperative}. However, most of these studies focus on ideal coordination tasks with accurate state and trajectory information. In uncertain environments, these methods become ineffective due to various sources of uncertainty and interference.

Only a few works aim to develop robust coordination strategies in realistic and complex scenarios. 
For example, the work \cite{pan2023hierarchical} obtains the velocity trajectories of CAVs using tube-based model predictive control with robust invariant sets to accommodate vehicle modeling and sensor noises. More recently, \cite{vitale2022autonomous,vitale2022optimizing} characterize vehicle state uncertainties as ellipses and compute an acceleration profile using integer linear programming, which ensures the ellipses do not intersect with each other under the assumption of not accounting for the vehicles' approaching direction. It is pertinent to note that these studies adopt inflexible maneuvers, such as prohibiting turns and restricting CAVs to predefined fixed paths, limiting the ability to avoid local obstacles. \cite{zhou2021distributed} accounts for tracking errors and deviations from nominal trajectories, and proposes a convex feasible set-based replanning coordination approach that iteratively solves a sequence of transformed subproblems within the convex sets. Further, to provide finer-grained safety modeling for collision avoidance under control and localization noise, the authors in\cite{chen2022re,li2024tightly} employ variable safety redundancy and search non-overlapping spatiotemporal resource blocks for vehicles. 
However, a major shortcoming in these studies is that they do not directly consider the evolution of trajectory uncertainties. During vehicle movement, trajectories tend toward deviations and divergence due to the accumulation of various uncertainties, which markedly increases the risk of interaction collisions.
Moreover, to simplify the complexity of safety conditions under uncertainties, these studies employ simple and fixed safety constraints based on vehicle and collision area size, which often fail to accommodate uncertain scenarios and ensure effective safety. 

Furthermore, real-world scenarios involving dynamic  multi-directional vehicle flows typically introduce high complexity and non-convexity into the optimization problem, making real-time solutions challenging, especially for larger numbers of CAVs.
Nowadays, numerical optimization tools have been well developed\cite{boyd2011distributed,yang2019survey}. As an emerging technique with scalable and parallel characteristics, the alternating direction method of multipliers (ADMM) can be well-suited to address the cooperative trajectory planning problem for CAVs\cite{saravanos2023distributed,zhang2021semi,cheng2024alternating,huang2023decentralized}.
The basic idea of ADMM is to decompose the original optimization problem into manageable subproblems and solve them separately and in parallel, thereby alleviating the computational burden arising from the growth in system dimensions.
\cite{zhang2021semi,cheng2024alternating} propose a cooperative trajectory planning algorithm based on ADMM to separate the independent constraints from the coupling constraints of the multi-agent system, allowing the former ones to be handled in parallel, thus resulting in a partially parallel solution.
Further, leveraging the ADMM, the authors in\cite{huang2023decentralized} mathematically formulate a consensus optimization problem over a connected undirected graph and then effectively distribute the computational load across all participating CAVs.
However, collision avoidance requirements in distributed frameworks result in a significant increase in the number of coupling constraints among agents. Therefore, these approaches often suffer from substantial computational burdens and conservative coordination performance. To realize real-time implementation, it remains an open issue to further improve computational efficiency through advanced ADMM-based optimization techniques. 

In summary, existing coordination methods often suffer from limited uncertainty treatment in safety modeling, where deterministic state assumptions or fixed safety margins cannot reliably capture uncertainty accumulation and interaction risks, and poor scalability and real-time performance under complex multi-vehicle coupling, where dense interactions render the resulting optimization highly non-convex and computationally expensive as the number of CAVs increases. As a result, ensuring reliable safety in dynamic uncertain environments without sacrificing computational efficiency remains a fundamental tension.

\subsection{Contributions}

This paper proposes a robust and real-time coordination framework for cooperative trajectory planning of CAVs that directly addresses the twin challenges of safety under uncertainty and computational efficiency. This work makes the following key contributions:

\begin{itemize}
    \item Novel uncertainty-aware safety formulation: We develop a robust multi-vehicle coordination framework that plans the nominal trajectory while directly controlling trajectory distribution evolution during vehicle interactions. Our formulation incorporates enhanced collision avoidance chance constraints with adaptive safety distances and terminal constraints that are expressed in terms of vehicle state mean and covariance. This formulation provides uncertainty-aware safety guarantees while dynamically adapting to time-varying scenarios.

    \item Computationally efficient distributed algorithm: We develop a fully parallel ADMM-based distributed trajectory negotiation (ADMM-DTN) algorithm that decomposes the complex optimization problem into manageable distributed subproblems. Our algorithm features two key innovations: 1) a negotiation process based on Jacobi updates that guarantees feasible solutions for any number of negotiation rounds, and 2) an interactive attention mechanism that identifies and prioritizes critical vehicle interactions, thereby significantly reducing computational overhead and coupling scale and enabling real-time deployment with varying computational resources.

    \item Comprehensive validation: We demonstrate the practical effectiveness of our framework through extensive simulations and real-world experiments. Our approach achieves up to 29.86\% and 40.79\% reductions in collision rates compared to benchmark methods while maintaining real-time performance and strong scalability with increasing vehicle numbers. Notably, practical experiments further validate real-time feasibility and robustness in the presence of environmental disturbances and unexpected non-CAV dynamic obstacles, addressing limitations common in existing coordination methods.
\end{itemize}

The proposed framework bridges the gap between theoretical guarantees and practical implementation in CAV coordination, offering a scalable solution for safe and efficient autonomous driving in complex traffic scenarios.

The paper is organized as follows. Section~\ref{section II} presents the intersection model, the vehicle model, and the formulation of the cooperative vehicle coordination problem. Section~\ref{section III} presents the enhanced safety constraints formulation and the robust multi-vehicle covariance steering model predictive control reformulation. Then, Section~\ref{sectionIV} presents a fully parallel ADMM-based distributed trajectory negotiation algorithm. Extensive simulations and practical experiments are presented in Sections~\ref{sectionV} and~\ref{sectionVI} to confirm the robustness and effectiveness of the proposed framework. Finally, Section~\ref{sectionVII} presents the conclusion and future work.

\emph{Notation}: Throughout this article, $\bar{x}$, $\hat{x}$ denote the mean and estimate of vector $x$. $P$, $\hat{P}$ denote the corresponding covariances.
The superscript $(\cdot)^k$ indicates the value at time step $k$. The subscript $(\cdot)_i$ indicates vehicle $i$. $\|x\|$ denotes the Euclidean norm of vector $x$. $\|x\|_Q^2=x^T Q x$ denotes the weighted 2-norm with appropriate dimensions of $x$ and matrix $Q$.
$\mathbb{R}^{n}$ and $\mathbb{R}^{m \times n}$ denote the set of $n$-dimensional real vectors and $m$ × $n$ real matrices, respectively. The set of integers $\mathbb{I}_{a: b}$ is defined by $\{a, a+1,...,b\}$. $\succeq$ denotes positive semidefinite. 
$\mathbb{E}$[·] denotes the expectation operation. $\operatorname{Pr}$[·] indicates the probability of an event and $\mathbb{P}$[·]  indicates the probability density function. 

\section{Problem Statement}\label{section II}
\subsection{Intersection Model}

\begin{figure}[htbp]
	\centering
	\begin{subfigure}[t]{0.665\columnwidth}
		\centering
		\includegraphics[width=\textwidth,trim=48 11 56 18.5,clip,valign=c]{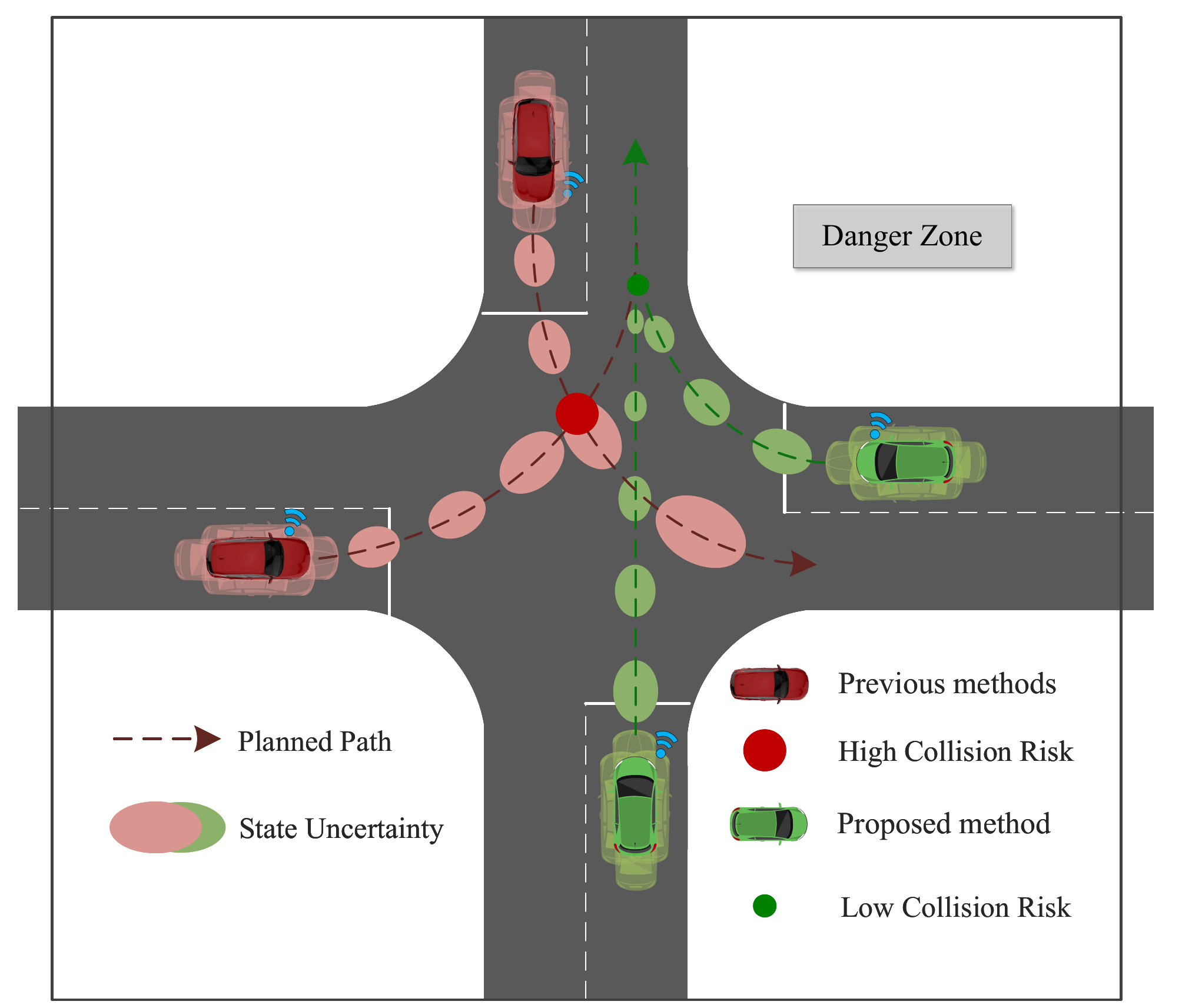}
		\vspace{-0.2em}
		\caption{} \label{1a}
	\end{subfigure}
	\hfill
	\begin{subfigure}[t]{0.32\columnwidth}
		\centering
		\includegraphics[width=\textwidth,trim=117 11.5 9.5 11.5,clip,valign=c]{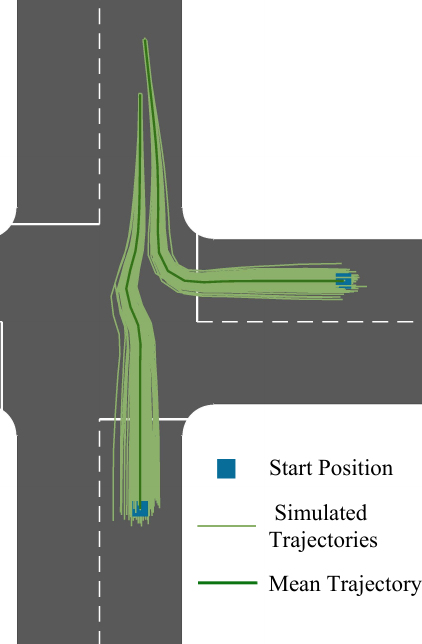}
		\vspace{1.3em}
		\caption{} \label{1b}
	\end{subfigure}
	\caption{An illustration of vehicle coordination under uncertainties. (a) Accumulating and expanding state uncertainties (pink), and shrinking state uncertainties (green) result in distinctly high and low interactive collision risks. (b) robust interaction of all simulated trajectories with converging trajectory distribution evolution, ensuring interaction safety.}
	\label{Fig01}
\end{figure}

We consider a typical unsignalized intersection which consists of $R$ roads with lane width $\mathcal{W}_{\text {lane }}$ (usually $R = 4$), as illustrated in Fig.~\ref{Fig01}. The area around the intersection is called the \emph{danger zone}. A fleet of vehicles passes through the intersection. CAVs move along flexible paths rather than being restricted to predefined fixed paths to improve road space utilization and flexibly respond to local obstacles.

The uncertainty in coordination stems from motion uncertainty, which arises from factors such as model inaccuracies and external disturbances. Moreover, sensor noises further corrupt the measurements. As a result, only an imprecise approximation of CAVs’ present and future states can be obtained. 
In previous methods, as illustrated by expanding pink ellipses in Fig.~\ref{Fig01}(\subref{1a}), the 3$\sigma$ error of vehicle state uncertainty gradually accumulates during vehicle movement. Consequently, vehicle trajectories tend to deviate and disperse from planned trajectories, leading to high interactive collision risks.
In our proposed method, as illustrated by the shrinking green ellipses in Fig.~\ref{Fig01}(\subref{1a}), the state uncertainty is steered, and trajectory distribution evolution is directly controlled, ensuring both convergence and robust interactions of the planned trajectories during vehicle interactions, as shown by the Monte Carlo trajectories in Fig.~\ref{Fig01}(\subref{1b}). The specific mechanism for mitigating the uncertainties is detailed in Lemma 1.

Moreover, CAVs can share driving information (e.g., position, speed, heading angle, and intention) with others through wireless communication. The information topology of multiple CAVs can be represented by an undirected graph $\mathcal{G} = (\mathcal{V}, \mathcal{E})$, where $\mathcal{V}=\left\{1,2, \ldots, M\right\}$ denotes the set of vehicles, and $\mathcal{E}=\left\{1,2, \ldots, m\right\}$ denotes the set of communication links between two interconnected vehicles, which can be defined as
\begin{equation}
\begin{cases}(i, j) \in \mathcal{E}(t), & \left\|p_i-p_j\right\| \leq d_{\mathrm{cmu}} \\ (i, j) \notin \mathcal{E}(t), & \text { otherwise },\end{cases}
\end{equation}
where $d_{\mathrm{cmu}}$ represents the communication range limit; $p_i$, $p_j$ are the position vectors of CAV $i$ and $j$. 
Based on the communication topology, the interactive neighbor set $\mathcal{N}_i$ of each CAV can be defined to construct collision avoidance relations. Meanwhile, the neighbor set $\mathcal{N}_i$ can be further refined by the interactive attention mechanism (discussed in Section~\ref{section IVB}) to selectively focus on critical interactive vehicles. 

\subsection{Vehicle Model}
Each CAV $i \in \mathcal{V}$ is subject to the following discrete-time, stochastic and nonlinear dynamics 
\begin{equation}\label{oridynamic}
x_i^{k+1}=f(x_i^k, u_i^k, w_i^k),
\end{equation}
for $k=0, \ldots, N$, where $x_i^k = [\mathrm{x}_i^k,\mathrm{y}_i^k,{\theta}_i^k,\mathrm{v}_i^k ]^{\top}\in \mathcal{X}_i \subset \mathbb{R}^{n_x}$ denotes the state of the CAV (position, orientation and velocity) and $u_i^k\in \mathcal{U}_i = [a_i^k,\delta_i^k]^{\top} \subset \mathbb{R}^{n_u}$ is the control input (acceleration and steering angle). $\mathcal{X}_i$ and $\mathcal{U}_i$ are state space and control space respectively. The increments of the process noise $w_i^k \in \mathbb{R}^{n_w}$ are the i.i.d. standard Gaussian random vectors. In this work, we adopt the nonlinear bicycle kinematic model as the coordination-layer motion model, focusing on safe interactive planning under uncertainty \cite{luo2023real,ge2021numerically}.
Residual model mismatch and external disturbances (e.g., unmodeled tire/actuator dynamics, tire–road friction, and environmental disturbances) are captured by stochastic disturbance and noise terms and are handled through the uncertainty-aware formulation presented in Section \ref{section III}.
The position of each vehicle in 2D space is extracted by $p_i^k=\Gamma x_i^k\in \mathbb{R}^{n_p}$, with $\Gamma \in \mathbb{R}^{n_p \times n_x}$ defined accordingly.

Furthermore, the states of vehicles are observed by the noisy and partial sensing model
\begin{equation}\label{measurement}
y_i^k=h(x_i^k, v_i^k),
\end{equation}
where $y_i^k \in \mathbb{R}^{n_y}$ is the measurement state and $v_i^k \in  \mathbb{R}^{n_y}$ is the measurement noise which is the i.i.d. standard Gaussian random vector. The noise processes $w_i^{0:N-1}$ and $v_i^{0:N-1}$ are independent. We defined the estimated state as $\hat{x}_i^k=\mathbb{E}\left(x_i^k \mid \mathscr{F}_i^k\right)$ and the prior estimated state as $\hat{x}_i^{k^{-}}=\mathbb{E}\left(x_i^k \mid \mathscr{F}_i^{k-1}\right)$, and the corresponding estimation errors are denoted as $\tilde{x}_i^k=x_i^k-\hat{x}_i^k$ and $\tilde{x}_i^{k^{-}}=x_i^k-\hat{x}_i^{k^{-}}$.

\subsection{Problem Formulation}
We consider a robust vehicle coordination task that cooperatively plans and controls $M$ CAVs through the intersection. Each CAV $i$ enters the intersection with its initial condition $x_i^0$, which is drawn from a multivariate Gaussian distribution $x_i^0 \sim \mathcal{N}\big(\mu_i^0, P_i^0\big)$, where $\mu_i^0 \in \mathbb{R}^{n_x}$ is the mean and $P_i^0 \succeq 0 \in \mathbb{R}^{n_x \times n_x}$ is the covariance matrix, respectively. 

The collision avoidance constraint is considered for safe interaction. In this work, vehicles are represented by rectangles, with $L_\text{c}$ and $W_\text{c}$ denoting the length and width of vehicles, respectively. Since the safety distances at the front and rear of the vehicle are larger, while the safety distances at the sides are smaller, the collision regions can be appropriately denoted as ellipses. We define the collision-checking region of vehicle $i$ with respect to vehicle $j$ by augmenting its length and width with the longitudinal size of vehicle $j$. This enables collision detection at any relative orientation while maintaining strict and rigorous discrimination. Thus, the collision condition of vehicle $i$ with vehicle $j$ at time $k$ is defined as:
\begin{equation}
\mathcal{C}_{i j}^k:=\left\{x_i^k \; \Big| \big\|p_i^k-p_j^k\big\|_{\Omega_{ij}}^2 \leq d_\text{s}\right\},
\end{equation}
where $\Omega_{i j}\!=\!R_o^T \!\operatorname{diag}\!\big(1 /\!\left(L_\text{ci}/2+\!L_\text{cj}/2\right)^2\!\!, 1 /\!\left(W_\text{ci}/2+\!L_\text{cj}/2\right)^2\!\big)\allowbreak  R_o$ and $d_\text{s} \in \mathbb{R}$ is normalized safety distance between the vehicle $i$ and vehicle $j$, typically equal 1. It is automatically tuned via $l_1$-regularization in Section~\ref{section IIIA}. Given that the states of the vehicles are random variables, we employ the collision avoidance chance constraints instead of hard constraints. Thus, based on the collision condition, the probability of the collision-free state is enforced to exceed a pre-specified threshold, i.e.,
\begin{equation}\label{avoidc}
\operatorname{Pr}\left(x_i^k\notin \mathcal{C}_{ij}^k\right) \geq 1-\xi_{\text {coll }}^x, \quad i \in \mathcal{V}, j \in \mathcal{N}_i,
\end{equation}
where $\xi_{\text {coll }}^x \in(0,0.5)$ is the maximum allowed probability of inter-vehicle collision. To enhance robustness to various uncertainties, we aim to drive the vehicles' final state distributions to target distributions $x_i^f \sim \mathcal{N}\big(\mu_i^f, P_i^f\big)$, where $P_i^f \succ 0 $.

Let $u=\big\{(u_i^k)_{k=0}^{N-1}, ..., (u_M^k)_{k=0}^{N-1}\big\}$ denote the control sequences of all vehicles, the cost function is given by
\begin{equation}\label{obf}
J(u)=\mathbb{E}\left[\sum_{i=1}^M\! \sum_{k=0}^{N-1}\!\left\|x_i^k-x_{i,r}^k\right\|_{Q_i^k}^2+\left\|u_i^k\right\|_{R_i^k}^2\right],
\end{equation}
where $N$ is short horizon, $Q_i^k \succeq 0$, $R_i^k  \succ 0$ are the weighting matrices, and $x_{i,r}^k$ is the desired state. The first term of the cost function penalizes the deviation of the vehicle state $x_i^k$ from the corresponding desired state $x_{i,r}^k$, and the second term penalizes the magnitude of the control input $u_i^k$. Specifically, the terminal cost is explicitly replaced by constraints on the mean and covariance of the target state. Here, the desired state vector $x_{i,r}^k$ is sampled at the maximum speed. Thus, the efficiency of the coordination will be ensured. 

As a result, at any time instant $t$, the centralized multi-vehicle coordination task can be formulated as the finite horizon stochastic optimization problem as follows
\begin{subequations}
	\begin{align}
	\mathscr{P}_1\!:  &\min_{x_i^k,u_i^k}\! \mathbb{E}\left[\sum_{i=1}^M \!\sum_{k=0}^{N-1}\!\left\|x_i^k-x_{i,r}^k\right\|_{Q_i^k}^2\!\!+\left\|u_i^k\right\|_{R_i^k}^2\right] \\
	\text { s.t. } & x_i^{k+1}=f(x_i^k, u_i^k, w_i^k), \!\!\! \quad k \in \mathbb{I}_{0: N-1} \\
	& y_i^k=h(x_i^k, v_i^k), \quad k \in \mathbb{I}_{0: N} \\
	& \operatorname{Pr}\!\big(x_i^k \notin \mathcal{C}_{ij}^k\big)\! \geq 1-\xi_{\text {coll }}^x, \!\!\!\quad i \in \mathcal{V}, j \in \mathcal{N}_i \\
	& x_i^N \sim \mathcal{N}\big(\mu_i^f, P_i^f\big) \label{termianl}\\
	& u_i^k  \in \mathcal{U}_i, \quad    k \in \mathbb{I}_{0: N-1} \\
	& x_i^k \in \mathcal{X}_i, \quad   k \in \mathbb{I}_{0: N}
	\end{align}
\end{subequations}
Similar to\cite{zheng2024cs,zhu2019chance}, since the prediction horizon is short and the control input at time $k$ is an affine function of the measurement data, it follows that the states will be Gaussian distribution over the horizon.
The challenges of solving the problem $\mathscr{P}_1$ in dynamic and uncertain environments are twofold.
The first challenge is to provide probabilistic guarantees for the interactions of stochastic state trajectories while ensuring highly robust coordination.
Second, multi-vehicle coordination problem is computationally hard to solve, as the computational complexity grows drastically with the number of vehicles, states, and time steps. 

In the following sections, we present a safe high-performance and computationally efficient multi-vehicle coordination framework.

\section{Uncertainty-Aware Coordination Safety Formulation}\label{section III}
In this section, to address the key coordination safety challenges, we reformulate the original stochastic, nonlinear, and nonconvex coordination problem into a robust cooperative planning and control problem. 
First, we address the intractable safety constraints, including collision avoidance chance constraints and terminal constraints, which comprehensively account for uncertainty levels and time-varying scenarios, and develop an adaptive and robust collision avoidance ability.
Then we reformulate the coordination problem into a multi-vehicle covariance steering model predictive control problem that quantify and steer the first and second moments of the vehicle state to ensure robust interactions among stochastic state trajectories.

\subsection{Enhanced Safety Constraints Formulation}\label{section IIIA}
\begin{figure}[t]
	\centering
	\begin{subfigure}{0.45\columnwidth}
		\centering
		\includegraphics[width=\textwidth]{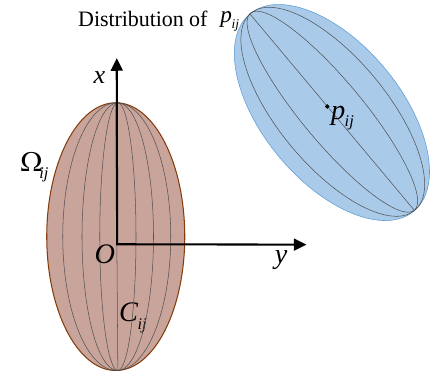}
        \vspace{-1.5em}
		\caption{}	\label{linearization_origin}
		\vspace{-0.5em}
	\end{subfigure}
	\begin{subfigure}{0.45\columnwidth}
		\centering
		\includegraphics[width=\textwidth]{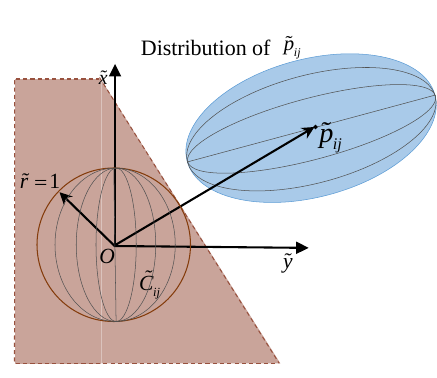}
        \vspace{-1.5em}
		\caption{} \label{linearized}
		\vspace{-0.5em}
	\end{subfigure}	
	\caption{Collision avoidance chance constraints reformulation. (a) Collision constraint with an ellipse region. (b) Transformation into a unit circle region and linearization.}
	\label{linearizations}
\end{figure}

First, we consider the inter-vehicle collision avoidance chance constraint. The position and uncertainty covariances of the two vehicles are extracted from the state variables and denoted by $p_i^k \sim \mathcal{N}\left(\bar{p}_i^k, P_i^k\right)$, $p_j^k \sim \mathcal{N}\left(\bar{p}_j^k, P_j^k\right)$, which follow multivariate Gaussian distributions. Then $p_i^k-p_j^k$ is also a multivariate Gaussian distribution, i.e. $p_i^k-p_j^k \sim \mathcal{N}\left(\bar{p}_i^k-\bar{p}_j^k, P_i^k+P_j^k\right)$. Therefore, the instantaneous collision probability of inter-vehicle can be expressed as a double integral of a multivariate Gaussian probability density function
\begin{equation}\label{integral}
\operatorname{Pr}\left(x_i^k \in \mathcal{C}_{ij}^k\right)\!=\!\!\int_{\left\|p_i^k-p_j^k\right\|_{\Omega_{ij}} \leq d_\text{s}} \!\!\! \mathbb{P}\left(p_i^k-p_j^k\right) d\left(p_i^k-p_j^k\right),
\end{equation}
where the integral region $\Omega_{ij}$, i.e., the collision region $\mathcal{C}_{i j}^k$ is an ellipse, as illustrated in Fig.~\ref{linearizations}(\subref{linearization_origin}). However, there is no analytical form to express the collision probability.
To obtain a tight upper bound, we first convert the collision region $\mathcal{C}_{i j}^k$ into a circle $\tilde{\mathcal{C}}_{i j}^k$ by employing an affine coordinate transformation $\tilde{x}=\Omega_{i j}^{\frac{1}{2}} x$, as shown in Fig.~\ref{linearizations}(\subref{linearized}). Thus the vehicle positions are transformed into new Gaussian distribution, i.e. $\tilde{p}_i^k \sim \mathcal{N}\big(\bar{\tilde{p}}_i^k, \tilde{P}_i^k\big), \tilde{p}_j^k \sim \mathcal{N}\big(\bar{\tilde{p}}_j^k, \tilde{P}_j^k\big)$, where
\begin{equation}
\bar{\tilde{p}}_i^k=\Omega_{i j}^{\frac{1}{2}} p_i^k, \quad \tilde{P}_i^k=\Omega_{i j}^{\frac{1}{2} {\top}} \!\! P_i^k \Omega_{i j}^{\frac{1}{2}},
\end{equation}
In the new coordinate system, the collision probability (\ref{integral})  is equal to the integral of $\tilde{p}_i^k-\tilde{p}_j^k$ over the circle $\tilde{\mathcal{C}}_{i j}^k$. Second, we approximate the circular collision region $\tilde{\mathcal{C}}_{i j}^k$ with a half space $\bar{\mathcal{C}}_{i j}^k$, which is denoted as
\begin{equation}
\bar{\mathcal{C}}_{i j}^k:=\left\{x \mid \alpha_{i j}^{k^{\top}}\left(\tilde{p}_i^k-\tilde{p}_j^k\right) \leq d_\text{s}\right\},
\end{equation}
where $\alpha_{i j}^{k}=\left(\bar{\tilde{p}}_i^k-\bar{\tilde{p}}_j^k\right) /\left\|\bar{\tilde{p}}_i^k-\bar{\tilde{p}}_j^k\right\|$. 
By linearizing the collision condition, it is evident that $\tilde{\mathcal{C}}_{i j}^k \subset \bar{\mathcal{C}}_{i j}^k$, thus $\operatorname{Pr}\left(x_i^k\! \in \mathcal{C}_{i j}^k\right) = \operatorname{Pr}\big(x_i^k\! \in \tilde{\mathcal{C}}_{i j}^k\big) \leq \operatorname{Pr}\left(x_i^k\! \in \bar{\mathcal{C}}_{i j}^k\right)$. Using the approach outlined in\cite{blackmore2011chance}, the collision probability upper bound between two vehicles is obtained:
\begin{equation}
\operatorname{Pr}\!\left(x_i^k \in C_{i j}^k\right)\!\leq\frac{1}{2}+\frac{1}{2} \operatorname{erf}\!\left(\!\frac{d_\text{s}-\alpha_{ij}^{k^{\top}}\!\left(\bar{\tilde{p}}_i^k-\bar{\tilde{p}}_j^k\right)}{\sqrt{2 \alpha_{i j}^{k^{\top}} \! \big(\tilde{P}_i^k+\tilde{P}_j^k\big)\alpha_{i j}^k}}\right)  ,
\end{equation}
where $\operatorname{erf}(x)$ is the standard error function, which is a monotonically increasing function. Given $\xi_{\text {coll }}^x \in(0,0.5)$,
thus, the probabilistic collision avoidance constraint (\ref{avoidc}) can be reformulated as a deterministic constraint,
\begin{equation}\label{Xchanceconstraint}
\begin{aligned}
\alpha_{ij}^{k^{\top}}\Omega_{i j}^{\frac{1}{2}}\big(\bar{p}_i^k-&\bar{p}_j^k\big)  -d_\text{s} \geq \operatorname{erf}^{-1}\!\!\left(1\!-2 \xi_{\text {coll }}^x\right)\\
&\cdot \sqrt{2 \alpha_{ij}^{k^{\top}}\Omega_{i j}^{\frac{1}{2} {\top}}\!\!\left(P_i^k+P_j^k\right)\Omega_{i j}^{\frac{1}{2}} \alpha_{i j}^k}.
\end{aligned}
\end{equation}

In practice, a static safety distance $d_{\text{s}}$ is inadequate for effective collision avoidance and struggles in highly dynamic intersection environments. To achieve time-varying collision avoidance ability, $d_{\text{s}}$ need to be adjusted flexibly. An intuitive mechanism would generate a larger $d_{\text{s}}$ for the upcoming time steps and a smaller $d_{\text{s}}$ for the farther time steps. Consequently, a dynamic safety distance approach is proposed, where a variable distance vector $d_i=(d_i^1,d_i^2,...,d_i^N)$ replaces $d_{\text{s}}$. Each $d_i^k$ varies between the bounds $[d_{\text{min}},d_{\text{max}}]$. However, this configuration slips the different importance of the upcoming and farther states, resulting in $d_i^k=d_{\text{min}}$ for all $k$. We wish to automatically allocate different attention to different states. To this end, the sparsity is imposed on $d_i$ to generate uneven values across different time steps. 
Mathematically, $l_1$-regularization is implemented by adding a penalty function of $d_i$ to the cost $J$. Let $d=\big\{(d_i^k)_{k=0}^{N}, ..., (d_M^k)_{k=0}^{N}\big\}$ denote the dynamic distance of all vehicles. This penalty is defined as the negative $l_1$ norm of $d_i$, i.e., $\tilde{J}(d)=-\eta\sum_{i=1}^M\|d_i\|_1=-\eta \sum_{i=1}^M\sum_{k=0}^N d_i^k$, where $\eta \in \mathbb{R}$ is a factor to adjust the collision avoidance ability.

Next, we relax the terminal constraint (\ref{termianl}) to the pair of convex constraints
\begin{equation}\label{terminalmean}
\mathbb{E}\left[x_i^N\right] \in \mathcal{X}_i^f,
\end{equation}
\begin{equation}\label{terminalcov}
\mathbb{E}\left[\big(x_i^N\!\!-\mathbb{E}\big[x_i^N\big]\big)\big(x_i^N\!\!-\mathbb{E}\big[x_i^N\big]\big)^{\top}\right] \preceq P_i^f,
\end{equation}
where $\mathcal{X}_i^f \subseteq \mathcal{X}_i \subset \mathbb{R}^{n_x}$ is compact and convex, and $P_i^f \in \mathbb{R}^{n_x \times n_x}$.

\subsection{Multivehicle Covariance Steering Model Predictive Control} 
For all vehicles, the trajectories are coordinated in a receding horizon fashion. Using the approach outlined in\cite{zheng2024cs}, the nonlinear system (\ref{oridynamic})-(\ref{measurement}) can be effectively approximated by linearization. Given a nominal trajectory generated from the previous iteration, successive linearization is employed to approach the nonlinear model more closely. Thus, the stochastic and linear time-varying dynamics is constructed as
\begin{equation}\label{linearization}
x_i^{k+1}=A_i^k x_i^k+B_i^k u_i^k+r_i^k+G_i^k w_i^k,
\end{equation}
\begin{equation}\label{mealinearization}
y_i^k=C_i^k x_i^k+D_i^k v_i^k,
\end{equation}
where $A_i^k \in \mathbb{R}^{n_x \times n_x}$, $B_i^k \in \mathbb{R}^{n_x \times n_u}$, $G_i^k \in \mathbb{R}^{n_x \times n_w}$ are the system matrices, and $C_i^k \in \mathbb{R}^{n_y \times n_x}$, $D_i^k \in \mathbb{R}^{n_y \times n_y}$ are the measurement matrices, and $r_i^k \in \mathbb{R}^{n_x}$ is the residual term computed from
\begin{align}
A_i^k \!\!=\!\!\left.\frac{\partial f}{\partial x}\right|_{\!\left(x_{i,r}^k, u_{i,r}^k, 0\right)}\!&,  
B_i^k \!\! =\!\!\left.\frac{\partial f}{\partial u}\right|_{\!\left(x_{i,r}^k, u_{i,r}^k, 0\right)}\!, 
G_i^k \!\!=\!\!\left.\frac{\partial f}{\partial w}\right|_{\!\left(x_{i,r}^k, u_{i,r}^k, 0\right)}\!, \nonumber \\
r_i^k =&f\left(x_{i,r}^k, u_{i,r}^k, 0\right)-A_i^k x_{i,r}^k-B_i^k u_{i,r}^k, \nonumber \\
C_i^k =&\left.\frac{\partial h}{\partial x}\right|_{\left(x_{i,r}^k, 0\right)},  D_i^k=\left.\frac{\partial h}{\partial v}\right|_{\left(x_{i,r}^k, 0\right)},
\end{align}

Since the true states of the CAVs are not available, the Kalman filter is used to estimate the states. By defining the innovation process $(\xi_i^{k})_{k=0}^N$ as 
\begin{equation}\label{innovation}
\xi_i^{k}=y_i^k-\mathbb{E}\left[y_i^k \mid \mathscr{F}_i^{k-1}\right]=C_i^k \tilde{x}_i^{k^{-}}+D_i^k v_i^k,
\end{equation}
where $\mathbb{E}\left(y_i^k \mid \mathscr{F}_i^{k-1}\right)=\mathbb{E}\left(C_i^k x_i^k+D_i^k v_i^k \mid \mathscr{F}_i^{k-1}\right)=C_i^k \hat{x}_i^{k^{-}}$. And $\xi_i^{k}$ is distributed as $\xi_i^{k} \sim \mathcal{N}\left(0, P_{\tilde{y}_{k^{-}}}\right)$. The innovation process at different time steps is uncorrelated as shown in\cite{aastrom2012introduction}. Hence, the covariance of each $\xi_i^{k}$ at time $k$ is 
\begin{equation}
P_{\xi_i^{k}}=\mathbb{E}\big(\xi_i^{k} \xi_i^{k^{\top}}\big)=C_i^k \tilde{P}_i^{k^{-}} C_i^{k^{\top}}+D_i^k D_i^{k^{\top}}.
\end{equation}
By substituting the innovation process (\ref{innovation}) and the Kalman priori filtered state into the posterior update, we obtain the estimated state dynamics of each CAV
\begin{equation}\label{estimation}
\hat{x}_i^{k+1}=A_i^k \hat{x}_i^k+B_i^k u_i^k+r_i^k+L_i^{k+1} \xi_i^{k+1},
\end{equation}
where $L_i^{k+1}$ is the Kalman gain. The state and measurement process (\ref{oridynamic}) and (\ref{measurement}) are fused and converted into a corresponding estimated state process (\ref{estimation}) with noise term $L_i^{k+1} \xi_i^{k+1}$. The coordination optimization problem can be posed in terms of the accessible estimated state.

To address the problem $\mathscr{P}_1$ robustly, we consider the following affine state feedback control policies
\begin{equation}\label{controllaw}
u_i^k=m_i^k+K_i^k\left(\hat{x}_i^k-\bar{x}_i^k\right),
\end{equation}
where $m_i^k \in \mathbb{R}^{n_u}$ is the feed-forward term that interprets as planning the mean of the filtered state, and $K_i^k \in \mathbb{R}^{n_u \times n_y}$ is the feedback gain that controls the system uncertainty. In each horizon, the first control command $u_i^0$ is executed. 

In this work, we choose the intermediate states of the planning horizon as decision variables, handling them in terms of their first and second moments. To this end, by substituting the control policy (\ref{controllaw}) into the estimated state dynamic (\ref{estimation}), then taking the expectation and computing the variance, we can obtain the mean and covariance dynamics of the vehicle state as follows
\begin{equation} \label{meandynamic}
\bar{x}_i^{k+1}= A_i^k \bar{x}_i^k + B_i^k m_i^k + r_i^k,
\end{equation}
\begin{equation}
\begin{aligned} \label{sigmadynamic}
\hat{P}_i^{k+1}\!=\!\left(A_i^k\!+\!B_i^k K_i^k\right)\!\hat{P}_i^k\!\left(A_i^k\!+\!B_i^k K_i^k\right)^{\top}\!\!\!\! + \! L_i^{k+1} P_{\xi_i^{k+1}} L_i^{{k+1 }^{\top}}.
\end{aligned} 
\end{equation}

Substituting $x_i^k=\hat{x}_i^k+\tilde{x}_i^k$ and using properties of conditional expectation, the state penalty of the cost function can be expressed in terms of the estimated state and the estimation error covariance. It is shown that the estimation error covariance is determined solely by the measurement model and thus is deterministic\cite{ridderhof2020chance}. Thus, this term can be discarded without changing the problem. Then, we rewrite the cost function in terms of the estimated state as
\begin{equation}\label{obf1}
\hat{J}(u)=\mathbb{E}\left[\sum_{i=1}^M\! \sum_{k=0}^{N-1}\!\left\|\hat{x}_i^k-x_{i,r}^k\right\|_{Q_i^k}^2+\left\|u_i^k\right\|_{R_i^k}^2\right].
\end{equation}
Similarly, the cost function can be further written in terms of the first two moments of the estimated state. Based on this controller structure and employing the transformation $U_i^k \triangleq K_i^k P_i^k$, the cost function (\ref{obf1}) can be converted into
\begin{equation}
\begin{aligned}\label{cost}
\hat{J}=  \sum_{i=1}^M \sum_{k=0}^{N-1}& \operatorname{tr}\big(Q_i^k \hat{P}_i^k\big)+(\bar{x}_i^k-x_{i,r}^k)^{\top} Q_i^k (\bar{x}_i^k-x_{i,r}^k) \\
& +\operatorname{tr}\big(R_i^k U_i^k \hat{P}_i^{k^{-1}} U_i^{k^{\top}}\big)+m_i^{k^{\top}} \! R_i^k m_i^k.
\end{aligned}
\end{equation}
Applying the same transformation process, the covariance dynamics (\ref{sigmadynamic}) becomes
\begin{equation}
\begin{aligned} \label{sigmadynamic1}
\hat{P}_i^{k+1}&=A_i^k \hat{P}_i^k A_i^{k^{\top}}  +B_i^k U_i^k A_i^{k^{\top} }+A_i^k U_i^{k^{\top}}  B_i^{k^{\top}} \\
&+B_i^k U_i^k \hat{P}_i^{k^{-1}} U_i^{k^{\top}} \!\!B_i^{k^{\top}}  +L_i^{k+1} P_{\xi_i^{k+1}} L_i^{{k+1 }^{\top}}.
\end{aligned}
\end{equation}

The problem remains non-convex due to the nonlinear term $U_i^k \hat{P}_i^{k^{-1}}\! U_i^{k^{\top}}$ appearing in the cost and covariance dynamics. To address this issue, we introduce the following relaxation 
\begin{equation} \label{relax}
U_i^k \hat{P}_i^{k^{-1}} U_i^{k^{\top}} -Y_i^k \preceq 0,
\end{equation}
the cost function (\ref{cost}) and constraint (\ref{sigmadynamic1}) can be rewritten as
\begin{equation}
\begin{aligned} \label{costrelax}
\hat{J}_i(\bar{x}_i^k,&m_i^k,\hat{P}_i^k,U_i^k,Y_i^k)=\sum_{k=0}^{N-1} \! \operatorname{tr}\big(Q_i^k \hat{P}_i^k\big)\!+\!\operatorname{tr}\left(R_i^k Y_i^k\right)\\
&+(\bar{x}_i^k-x_{i,r}^k)^{\top} Q_i^k (\bar{x}_i^k-x_{i,r}^k) 
+\!m_i^{k^{\top}} R_i^k m_i^k,
\end{aligned}
\end{equation}
\begin{equation}
\begin{aligned} \label{sigmadynamicrelax}
\hat{P}_i^{k+1}=&A_i^k \hat{P}_i^k A_i^{k^{\top} } +B_i^k U_i^k A_i^{k^{\top}} +A_i^k U_i^{k^{\top}}  B_i^{k^{\top}} \\
&+B_i^k Y_i^k B_i^{k^{\top}}  +L_i^{k+1} P_{\xi_i^{k+1}} L_i^{{k+1 }^{\top}}.
\end{aligned}
\end{equation}
This convex relaxation turns out to be lossless, see\cite{liu2024optimal} for details, where (\ref{relax}) can be further written as the linear matrix inequality (LMI) using the Schur complement. 
\begin{equation}\label{relaxation}
\left[\begin{array}{ll}\hat{P}_i^k & U_i^{k^{\top}} \\ U_i^k & Y_i^k\end{array}\right] \succeq 0.
\end{equation}
Using $P_i^k = \hat{P}_i^k + \tilde{P}_i^k$, the inter-vehicle collision avoidance constraint (\ref{Xchanceconstraint}) is rewritten in terms of the estimated state.

Furthermore, within each optimization horizon, the cost function (\ref{costrelax}) is minimized with respect to the initial state conditions of the vehicles ($\bar{x}_i^k, P_i^k$). The initial state involves selecting either the current measurement state or the predicted state.
We employ a dual initialization strategy similar to\cite{knaup2023safe}, which determines the initial conditions for each optimization horizon based on the optimality of the cost function. Specifically, the measurement state is used if the problem is feasible and leads to a lower cost than the problem with predicted state. Otherwise, the predicted state is selected.
 
In a nutshell, the robust cooperative planning and control problem is reformulated as follows
\begin{subequations}
\begin{align}
\mathscr{P}_2:  &\min _{u_i^k\in \Pi} \sum_{i=1}^M \hat{J}_i(\bar{x}_i^k,m_i^k,\hat{P}_i^k,U_i^k,Y_i^k) +\tilde{J}(d)  \\
\text { s.t. } & E_i^k(\bar{x}_i^{k+1}, \bar{x}_i^k, m_i^k)=0, \quad k \in \mathbb{I}_{0: N-1} \label{b} \\
& F_i^k(\hat{P}_i^{k+1},\hat{P}_i^k, U_i^k, Y_i^k)=0, \quad k \in \mathbb{I}_{0: N-1}  \label{c}\\
& H_i^k(\hat{P}_i^k, U_i^k, Y_i^k) \succeq 0, \quad k \in \mathbb{I}_{0: N} \label{d} \\
& I_{ij}^k\big(\bar{x}_i^k,\hat{P}_i^k,\bar{x}_j^k,\hat{P}_j^k,d_i\big) \succeq 0, \quad i \in \mathcal{V}, j \in \mathcal{N}_i  \label{Iij}\\
& d_i^k \in\left[d_{\min }, d_{\max }\right], \quad  k \in \mathbb{I}_{0: N}  \label{f}\\
& u_{\min } \preceq m_i^{k}  \preceq u_{\max }, \quad k \in \mathbb{I}_{0: N-1} \label{g}\\
& \bar{x}_i^N \in \mathcal{X}_i^f \in \mathcal{X}_i, \quad \bar{x}_i^k \in \mathcal{X}_i \label{h}\\
&\hat{P}_i^N \preceq P_i^f-\tilde{P}_i^N \label{i}
\end{align}
\end{subequations}
where $E_i^k$ and $F_i^k$ denote mean and covariance dynamics (\ref{meandynamic}), (\ref{sigmadynamicrelax}) respectively. $H_i^k$ denotes the relaxation constraint (\ref{relaxation}). $I_{ij}^k$ denotes the inter-vehicle collision avoidance constraint (\ref{Xchanceconstraint}).
At each time instant $t$, the reformulated coordination problem $\mathscr{P}_2$ aims to find the decision variables $\bar{x}_i,m_i,P_i,U_i,Y_i$ for all vehicles. Then the control policies can be recovered from the computed variables. 
To better explain the uncertainty-mitigation mechanism of the proposed framework, we next present a conditional analytical result on covariance attenuation and bounded uncertainty evolution.

\emph{Lemma 1}: Consider uncertainty-aware coordination under the affine feedback policy (\ref{controllaw}) and the corresponding mean/covariance propagation (\ref{meandynamic}), (\ref{sigmadynamic}). Given that the closed-loop covariance propagation is $\kappa$-contractive in the positive semidefinite order through the computed feedback gain $K_i^k$, i.e., $\left(A_i^k+B_i^k K_i^k\right) \hat{P}_i^k\left(A_i^k+B_i^k K_i^k\right)^{\top} \preceq \kappa^2 \hat{P}_i^k$. Then, if the initial covariance satisfies $\hat{P}_i^0 \succeq P_{\infty}$, then the covariance will decrease monotonically to the steady-state bound $P_{\infty}$:
\begin{equation}
\hat{P}_i^{k+1} \preceq \hat{P}_i^k \quad \text { whenever } \quad \hat{P}_i^k \succeq P_{\infty} . 
\end{equation}
        
Proof. See Appendix \ref{proof}.

\section{Parallel Trajectory Negotiation}\label{sectionIV}
To address key real-time computational challenges, we propose a completely parallel computation framework to efficiently solve the problem $\mathscr{P}_2$. We first present the ADMM-DTN algorithm to decompose the optimization problem into two manageable subproblems, which are solved in parallel across vehicles through trajectory negotiation. Then, we incorporate an interactive attention mechanism that selects the most critical interactive vehicles to further reduce computational burden.
For the sake of clarity, we omit the time step index $k$. 

\subsection{Parallel Computation Via ADMM-DTN}
\begin{figure}
	\centering
	\includegraphics[width=1\linewidth]{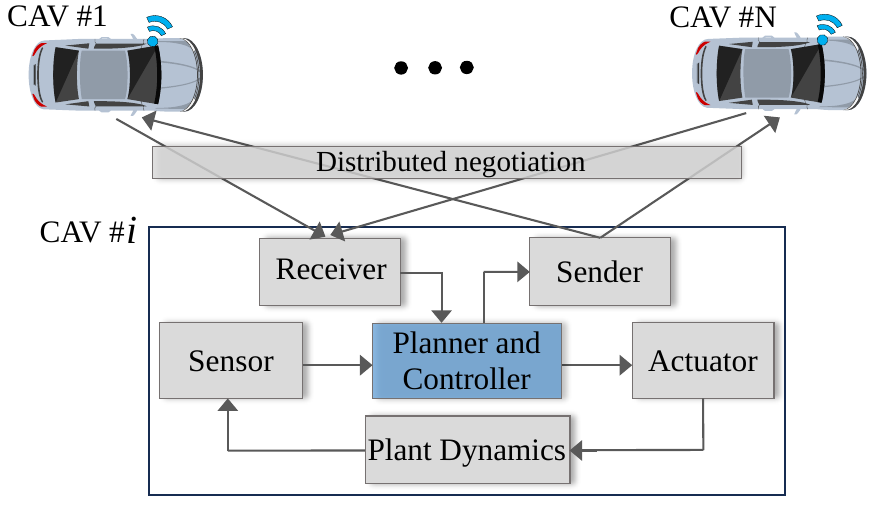}\\
	\vspace{-0.5em}
	\caption{Distributed trajectory computation and negotiation architecture.}\label{Fig03}
	\vspace{-1em}
\end{figure}
The collision avoidance constraints remain intractable as the variables $\bar{x}$ and $P$ are coupled within the square root. To address this issue and enable parallel computation, firstly, we decompose the problem $\mathscr{P}_2$ into two tractable subproblems associated with different sets of variables by leveraging the ADMM. In particular, the augmented Lagrangian function of problem $\mathscr{P}_2$ can be formulated as
\begin{align}\label{augmentionlagrangian}
 \mathcal{L}_\rho(\bar{x}&,m,d,\hat{P},U,Y,\Delta;\lambda) 
=\sum_{i=1}^M J_i(\bar{x}_{i},m_{i},\hat{P}_{i},U_{i},Y_{i})+\tilde{J}(d) \nonumber\\
&+\frac{\rho}{2} \sum_{i=1}^{M} \sum_{j=1}^{\mathcal{N}_i}\left\|I_{ij}\big(\bar{x}_{i},\hat{P}_{i},\bar{x}_{j},\hat{P}_{j},d_i,\Delta_{i j}\big)\!+\!\lambda_{i j}\right\|_2^2,
\end{align}
where $\rho$ is the penalty parameter, and $\lambda_{ij}$ is the dual variable corresponding to equality constraint $I_{ij}\big(\bar{x}_{i},\hat{P}_{i},\bar{x}_{j},\hat{P}_{j},d_i,\Delta_{i j}\big)=0$. The equality constraint is transformed from the constraint (\ref{Iij}), where the nonlinear function $I_{ij}\big(\bar{x}_{i},\hat{P}_{i},\bar{x}_{j},\hat{P}_{j},d_i,\Delta_{i j}\big)$ is given by
\begin{equation}
\begin{aligned}
&I_{ij}\big(\bar{x}_{i},\hat{P}_{i},\bar{x}_{j},\hat{P}_{j}, d_i, \Delta_{i j}\big)= (\alpha_{i j}^{\top}\!\left(\bar{p}_{i}-\bar{p}_{j}\right)-d_i)^2\\ 
&- 2 (\operatorname{erf}^{-1}\!\!\left(1\!-2 \xi_{\text {coll }}^x\right))^2 \alpha_{i j}^{\top}\big(\hat{P}_{i}+\hat{P}_{j}+\tilde{P}_{i}+\tilde{P}_{j}\big) \alpha_{i j} - \Delta_{i j},
\end{aligned}
\end{equation}
where $\Delta_{i j}$ is the slack variable. In (\ref{augmentionlagrangian}), the primal variables are divided into two groups: 1) $\{\bar{x},m,d\}$, which are associated with the mean dynamic; 2) $\{P,U,Y,\Delta\}$, which are related to the covariance dynamic. And updates are made to the dual variables $\lambda_{ij}$, which correspond to different inter-vehicle collision avoidance constraints. As a result, the ADMM algorithm for minimizing the augmented Lagrangian can be denoted by 
\begin{subequations}
	\begin{align}
     \bar{x}^{l+1}\!\!,m^{l+1}\!,\!\!\!\!\!\quad&d^{l+1}\!\!\!=\!\underset{\bar{x},m,d}{\operatorname{argmin}} \mathcal{L}_\rho \!\! \left(\bar{x},m,d,P^l\!,U^l\!\!,Y^l\!\!,\Delta^l ; \lambda^l\right) \label{xv}\\
	 P^{l+1}\!,U^{l+1}&\!,Y^{l+1}\!\!,\Delta^{l+1}\!\!= \
    \notag \\
    &\underset{P,U,Y,\Delta}{\operatorname{argmin}} \mathcal{L}_\rho \! \left(\bar{x}^{l+1}\!,m^{l+1}\!,d^{l+1}\!, P,U,Y,\Delta ; \lambda^l\right) \label{sigma}\\
	 \lambda_{i j}^{l+1}\!=\lambda_{i j}^l&+I_{ij}\left(\bar{x}_i^l,P_i^l,\bar{x}_j^l,P_j^l, d_i^l, \Delta_{ij}^l\right). \label{dualupdate}
	\end{align}
\end{subequations}
Subproblems (\ref{xv}) and (\ref{sigma}) are convex, and (\ref{dualupdate}) is the subgradient for updating the dual variables.
The termination criteria are given in terms of the primal residual and dual residual, i.e.,
\begin{subequations}\label{termination}
	\begin{align}
	& \sum_i \sum_j\left\|I_{ij}\left(\bar{x}_{i},P_{i},\bar{x}_{j},P_{j}, d_i, \Delta_{i j}\right)\right\|_2^2 \leq \epsilon^{\mathrm{pri}}, \label{terminationa}\\
	& \sum_i \sum_j\left\|\lambda_{ij}^{l+1}-\lambda_{ij}^l\right\|_2^2 \leq \epsilon^{\text {dual }}, \label{terminationb}
	\end{align}
\end{subequations}
where constants $\epsilon^{\mathrm{pri}} \geq 0 $ and $\epsilon^{\text {dual }} \geq 0 $ are the threshold for the termination criteria.

Secondly, we present a distributed trajectory negotiation (DTN) algorithm based on Jacobi updates to achieve efficient parallel computation. Fig.~\ref{Fig03} shows the distributed computation and negotiation architecture, encompassing inter-vehicle communication with on-board planning and control. The vehicles are interconnected and share the resulting information with neighboring vehicles. Based on the shared information, each vehicle simultaneously performs trajectory planning and control and updates its plan, after which the updated trajectory is broadcast again for the next negotiation round. 
Thus, the two subproblems (\ref{xv}), (\ref{sigma}) can be solved in parallel for each vehicle. 
The local optimization decision of each vehicle is denoted as $s_i =(\bar{x}_i^k,m_i^k,\hat{P}_i^k,U_i^k,Y_i^k,d_i^k,\Delta_{ij})$, which includes the variables related to the $i$-th vehicle subsystem. This decoupled decomposition is practical and preserves the privacy of vehicle models\cite{doan2017jacobi,kneissl2019one}.
Each vehicle determines the local trajectory and the control input by computing $s_i$. To simplify notation, we state the local optimization problems as
\begin{subequations}\label{local}
	\begin{align}
	s_i^*=&\underset{s_i}{\operatorname{argmin}}  J_i\left(s_i\right) +\frac{\rho}{2} \sum_{j=1}^{\mathcal{N}_i} \left\|\mathcal{A}_{i j}^d s_i+\mathcal{B}_{i j}^d s_j-c_{i j}^d\right\|_2^2 \label{locala}\\
	&\text { s.t. }  \mathcal{A}_i s_i-b_i \leq 0  \label{localb}
	\end{align}
\end{subequations}
where $J_i\!\left(s_i\right)$ corresponds to the cost $J_i(\bar{x}_{i},m_{i},P_{i},U_{i},Y_{i})$ and $\tilde{J}_i(d_i)$, and $\mathcal{A}_{i j}^d s_i+\mathcal{B}_{i j}^d s_j-b_{i j}^d$ corresponds to the coupling collision avoidance constraint. For the mean subproblem (\ref{xv}), constraints (\ref{b}), (\ref{Iij})-(\ref{h}) are collected in (\ref{localb}), while for the covariance subproblem (\ref{sigma}), constraints (\ref{c})-(\ref{Iij}) and (\ref{i}) are collected in (\ref{localb}).

The procedure of the DTN is presented in Algorithm~\ref{Algorithm 1}, to improve the quality of the planned trajectory and mitigate the conservatism of parallel computation, the DTN algorithm negotiates among vehicles to reach a solution, where each vehicle $i \in \mathcal{V}$ perform the optimization computation and negotiation update in lines 3 and 5 iteratively.
For each individual vehicle, optimized trajectory information is modified through a local Jacobi update step (\ref{jacobupdate}) and shared with interactive neighbor vehicles through communication. 
The process can be interrupted after each negotiation while guaranteeing a feasible solution. Where $\dot{s}_i$ is an initial candidate at the start of negotiation procedure, $s^{(p)}_i$ is the feasible output of the $p$-th negotiation, and $s^{*}_i$ is the optimal solution of the local optimization problem. And the update variable $\omega_i$ sets the degree of over-relaxation. The negotiation process stops when either it reaches the maximum negotiation number $p_{\max}$ or the result $s_i$ converges to a specific stopping boundary $\gamma$. 

The DTN algorithm has two properties that ensure the effectiveness of parallel computation.
First, the output $s^{(p)}_i$ from each negotiation is feasible for local problems (\ref{local}), meaning that the negotiation process can be interrupted after any round, making it suitable for real-world deployment.
This is because, beginning with a feasible initial candidate, we select any vehicle $i$ and its any interacting neighbor vehicle $j \in \mathcal{N}_i$. The first negotiation is reorganized as (\ref{feasible}). Two compact solution vectors $\binom{s_i^*}{s_j^{(0)}}$ and $\binom{s_i^{(0)}}{s_j^*}$ are feasible for local problems of vehicles $i$ and $j$. Since the constraints and cost of the problems are convex, the convex combination (\ref{jacobupdate}) with $\omega_i + \omega_j = 1$ yields a new feasible solution $\binom{s_i^{(1)}}{s_j^{(1)}}$ for the same problems. The same result holds for $p >1$ by induction. 
Second, the cost over consecutive negotiation updates is non-increasing and converges as $p \rightarrow \infty$, thereby improving the solution quality.
This is because, using the convexity of the cost and the fact that $ s_i^*$ is the optimal solution of the $p$-th negotiation, we have $V_i\big(s_i^{(p+1)}\big)  =V_i\big(\omega_i s_i^*+\left(1-\omega_i\right) s_i^{(p)}\big) \leq \omega_i V_i\big(s_i^{(p)}\big)+\left(1-\omega_i\right) V_i\big(s_i^{(p)}\big)  =V_i\big(s_i^{(p)}\big)$. Thus, the cost sequence is bounded below and monotonic, ensuring convergence as $p \rightarrow \infty$. 
\begin{equation}\label{feasible}
\binom{s_i^{(1)}}{s_j^{(1)}}\! =\!\binom{\omega_i s_i^*+\left(1-\omega_i\right) s_i^{(0)}}{\omega_j s_j^*+\left(1-\omega_j\right) s_j^{(0)}} \! =\!\omega_i\binom{s_i^*}{s_j^{(0)}}\!+\!\omega_j\binom{s_i^{(0)}}{s_j^*}.
\end{equation}

\begin{algorithm}
	\small
	\caption{Distributed Trajectory Negotiation}
	\label{Algorithm 1}
	\LinesNumbered
	Initialization: current time $t$,  negotiation round $p$:= $0$,  each CAV $i \in V$ receive $s^{(p)}_j = \dot{s}_j, j \in \mathcal{N}_i$\;
	\For {negotiation round $p = 1, 2, \ldots$}{		
		compute $s^{*}_i \Big(\{ s^{(p)}_j \}_{j \in \mathcal{N}_i}\Big)$ in parallel for each $i \in V$\;
		determine modified trajectory with $\omega_i \! \geq \!0, \omega_i + \omega_j \!=\! 1$\;
		\begin{equation}\label{jacobupdate}
		s^{(p+1)}_i = \omega_i s^{*}_i \left(\{ s^{(p)}_j \}_{j \in \mathcal{N}_i}\right) + (1 - \omega_i)s^{(p)}_i 
		\end{equation}
		\nl share $s^{(p+1)}_i$ with CAVs $j \in \mathcal{N}_i$\;
		\If{$p > p_{\max}$ or $\left| s^{(p-1)}_i - s^{(p)}_i \right| < \gamma \quad \forall i \in V$}{
			break\;}
	}	
\end{algorithm}
\subsection{Interactive Attention Mechanism}\label{section IVB}

In scenarios involving a large number of vehicles, vehicles are mostly influenced by their nearby vehicles rather than all others. Thus, it is essential for vehicles to focus on the most interactive vehicles that could potentially cause conflicts. Inspired by\cite{leurent2019social}, we present an interactive attention mechanism based on a self-attention architecture, which selects the most interactive vehicles to reduce coupling constraints and further enhance computational efficiency.

\begin{figure}
	\centering
	\includegraphics[width=0.9\linewidth]{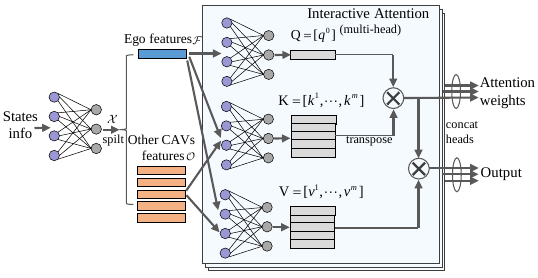}\\
	\vspace{-0.5em}
	\caption{Workflow of interactive attention mechanism for generating attention weights.}\label{Fig04}
	\vspace{-1em}
\end{figure}

The interactive attention mechanism is illustrated in Fig.~\ref{Fig04} and works as follows. 
The collected state information is embedded into the feature vectors $\mathcal{X} =\left(\mathcal{F}, \mathcal{O}\right) \in \mathbb{R}^{M \times d_x}$ through the linear layer, which contains the ego features $\mathcal{F}$ and other CAVs' features $\mathcal{O}$. The weights are shared among all vehicles. 
The difference is that driving intentions are also included in the feature information, facilitating faster learning of interactive relationships. 
Then the feature vectors $\mathcal{X}$ are fed into a multi-head attention layer and embedded into the query $\mathbf{Q}$, keys $\mathbf{K}$, values $\mathbf{V}$ through multilayer perceptrons (MLPs). 
\begin{equation}
 \mathbf{Q}=W^Q \mathcal{X}, \quad \! \mathbf{K}=W^K \mathcal{X}, \quad \! \mathbf{V}=W^V \mathcal{X},
\end{equation}
where $W^Q$, $W^K$, and $W^V \in \mathbb{R}^{d_x \times d_k}$. Unlike the attention layer in the Transformer model, this module produces only the query results (i.e., attention weights) which represent the degree of attention to surrounding vehicles.

To select interactive vehicles based on the surroundings, the vehicle first emits a single query $\mathbf{Q}=\left[q^0\right] \in \mathbb{R}^{1 \times d_k}$, which is then compared to all of keys $\mathbf{K}=\left[k^1, \cdots, k^M\right] \in \mathbb{R}^{M \times d_k}$, including the descriptive features of each vehicle. The similarity between the query $q^0$ and any vehicle's key $k^i$ is calculated using the dot product. These similarities are scaled by the inverse-square-root-dimension $\frac{1}{\sqrt{d_k}}$ and normalized with a softmax function $\sigma$ to obtain attention weights. Then, these weights gather all of the values $\mathbf{V}=\left[v^1, \cdots, v^M\right] \in \mathbb{R}^{M \times d_k}$, where each value $v^i$ is a feature computed through a shared linear projection. The attention computation for each head is expressed as 
\begin{equation}\label{attention}
\text { output }=\underbrace{\sigma\Big(\frac{\mathbf{Q K}^{\top}}{\sqrt{d_k}}\Big)}_{\text {attention weights }} \!\!\mathbf{V}.
\end{equation}
Finally, we combine the attention weights from all heads to represent the vehicle's attention to surrounding vehicles. 

\subsection{Proposed Algorithm} 
In outline, the proposed robust and real-time vehicle coordination framework is summarized in Algorithm~\ref{Algorithm 2}.
Specifically, the relative states of each vehicle and its interacting collision-avoidance vehicles do not change markedly between consecutive time steps. Warm starting can be leveraged to significantly reduce the number of iterations in ADMM. That is, the current solution can be used as the initial guess for the primal and dual variables in the next iteration procedure.
\begin{algorithm}
	\small
	\caption{The Robust and Real-Time Vehicle Coordination Framework}
	\label{Algorithm 2}
	\LinesNumbered
	\KwIn{CAV geometry $L_\text{c}$, $W_\text{c}$; communication network $\mathcal{G} = (\mathcal{V}, \mathcal{E})$; Motion and measurement noise matrix $G_i$, $D_i$; Horizon $N$; Probability thresholds $\xi_{\text {coll }}^x$.}
	Initialization: $t$:= $0$; Initial state of each CAV $\big(\bar{x}_i^{0}, \hat{P}_i^0\big)$; safety distance vector $d_i$; interactive neighbor set $\mathcal{N}_i$\;
	\While{\emph{(\ref{terminationa}) and (\ref{terminationb}) are not satisfied}}{
            Calculate attention weights by (\ref{attention}) to update the interactive neighbor set\;
		Update the variables $\bar{x}, m, d$ by solving (\ref{xv}) with Algorithm 1 in parallel for each $i \in \mathcal{V}$ \;
		Update the variables $P,U,Y,\Delta$ by solving (\ref{sigma}) with Algorithm 1 in parallel for each $i \in \mathcal{V}$ \;
		Update the Lagrangian multipliers $\lambda$ by (\ref{dualupdate}) in a parallel manner\;
	}	
	Obtain the control input by (\ref{controllaw}) and execute the first command\;
\end{algorithm}

\section{Simulation Results}\label{sectionV}
\subsection{Simulation Settings}\label{sectionVA}
    We implement extensive simulations to validate the performance of the proposed coordination framework, with all experiments conducted on a desktop computer with Intel i7-9700F 3.0 GHz CPU and 16 GB of RAM.
The considered intersection scenario is illustrated in Fig.~\ref{Fig01}. The road geometry is set as\cite{luo2023computationally}. The vehicle dynamics is discretized by 2nd-order Runge-Kutta integration, see the Appendix \ref{bicycle} for derivation details.
The probability threshold is set as $\xi_{\text {coll}}^x$ = 0.1 to provide CAVs a suitable safety level. 
The disturbance matrix is set as $G = \mathrm{diag}(0.08,0.08,\pi/180,0.1)$ based on the high-fidelity simulated driving data of the vehicle from\cite{knaup2023safe}. To imitate realistic scenarios, measurement noises are added to the recorded state, with the noise matrix set as $D = \mathrm{diag}(0.35,0.35,\pi/150,0.2)$ based on the level of localization error in real-world experiments\cite{li2018high,8795887}.
The penalty parameter $\rho$ is set as 100. The thresholds of termination criteria $\epsilon^{\mathrm{pri}}$ and $\epsilon^{\text {dual }}$ are set to 0.1. Given that the output of each negotiation is feasible, we set $p_{\text {max}} = 4 $ for subproblem (\ref{xv}) and $p_{\text {max}} = 2$ for subproblem (\ref{sigma}) to trade off computational efficiency and overall coordination performance. The update variable $w_i$ is set as 0.5.
The attention mechanism is deployed following the version specified by\cite{leurent2019social}, where the MLPs consist of two linear layers and the layer size of 64 × 64. The attention layer contains two heads, and the feature size is 32. The maximum number of interactive vehicles for each vehicle is set to 5.
The optimization frameworks CasADi and YALMIP, with the solvers IPOPT and MOSEK, respectively, are used to solve subproblems (\ref{xv}) and (\ref{sigma}). The detailed parameters are summarized in Table \ref{simupara}.

\setlength{\tabcolsep}{1.45mm}{
	\begin{table}[h]
		\vspace{-.5em} 
		\caption{Simulation Parameters}
		\centering
		\label{simupara}
		\begin{tabular}{cc}		
			\toprule[1pt]
			Parameters & Values \\
			\midrule
                size of intersection area              & 80 m $\times$ 80 m \\
			lane width                             & 10 m \\
			maximum velocity $v_{\text{max}}$      & 10 m/s    \\
			maximum acceleration $a_{\text{max}}$               &$[-5,5]$ m/s$^2$ \\  
			maximum steering angle $\delta_{\text{max}}$        &$[-0.78,0.78]$ rad  \\
			vehicle geometry $L_{\text{c}} / W_{\text{c}}$  &4.2 m/2.1 m \\
			vehicle	wheelbase $L_w$                & 3 m  \\ 
			weighting factor $\eta$                & 15  \\
			variable safety distance $d_{ij}  $    & [1.1-1.5]   \\
			time step $\tau$                       & 0.1 s     \\
			horizon length $N$                     & 20    \\
			state weight matrix $Q$                &$\operatorname{diag}(2,2,1,0)$ \\
			input weight matrix $R$                &$\operatorname{diag}(1,1)$ \\ 
			initial state covariance $\hat{P}_i^0$  & $\mathrm{diag}(0.4,0.4,\pi/90,0.2)$  \\ 
			prior estimation error covariance $\tilde{P}_i^{0^-}$  & $\mathrm{diag}(0.03,0.03,\pi/360,0.02)$ \\
                terminal state covariance $P_i^f$  & $\mathrm{diag}(0.15,0.15,\pi/180,0.1)$  \\ 
			measurement matrix $C$                 & $I_4$     \\     
			\bottomrule[1pt]
		\end{tabular}
\end{table}}

\subsection{Performance of the Proposed Framework}\label{sectionVB}

In this subsection, we conducted extensive comparative experiments, ranging from small-scale to large-scale, to validate the performance of the proposed coordination framework. 
We compare the proposed method with representative benchmarks in terms of coordination trajectory, safety, efficiency, and computational overhead.
The benchmarks cover two complementary coordination paradigms closely related to this study. 
The accelerated space-time resource searching (STRSv2) approach\cite{li2024tightly} represents a resource-block-based coordination paradigm. It emphasizes the efficient allocation of non-overlapping feasible tunnel and vehicle-level safety planning, retaining flexible adaptation to environmental uncertainty.
The convex feasible set-based distributed MPC (CFSDMPC) approach\cite{zhou2021distributed} represents a distributed optimization-based coordination paradigm. It accounts for trajectory deviation and tracking error and performs real-time multi-vehicle coordination by iteratively solving convexified subproblems.

\begin{figure}[htbp]
	\centering
	\begin{subfigure}{0.49\columnwidth}
		\centering
     	\includegraphics[width=\textwidth, angle=-90,trim=210 165 0 0,clip]{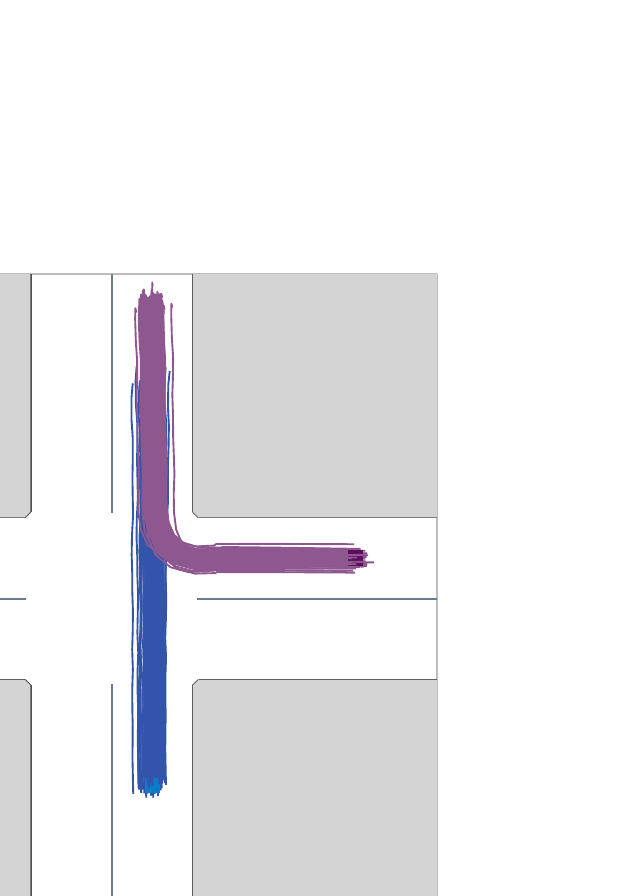}
		\vspace{-3.7em}
		\caption{STRSv2} 
		\vspace{0.5em}
	\end{subfigure}
	\centering
	\begin{subfigure}{0.49\columnwidth}
		\centering
		\includegraphics[width=\textwidth, angle=-90,trim=210 165 0 0,clip]{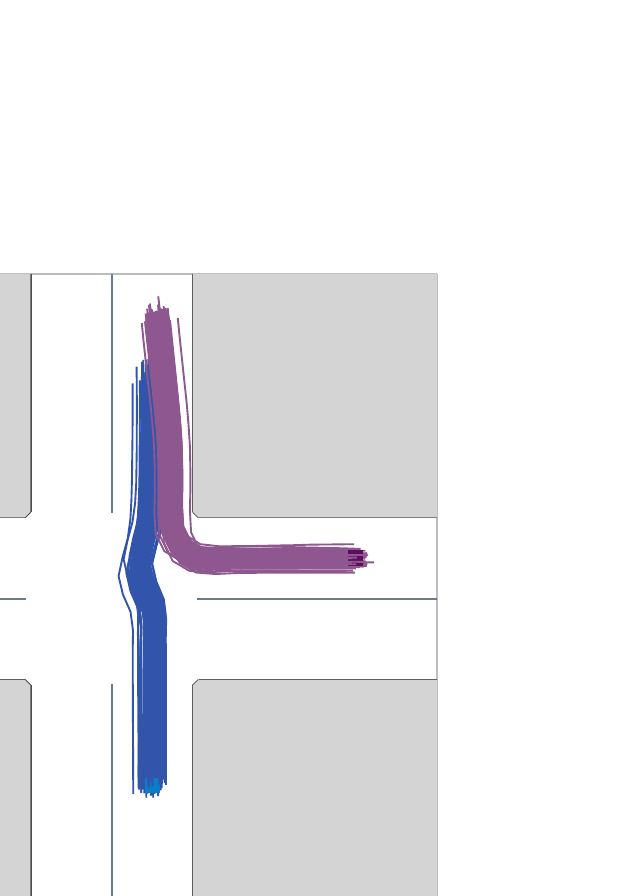}
		\vspace{-3.7em}
		\caption{CFSDMPC} 
		\vspace{0.5em}
	\end{subfigure}
	\centering
	\begin{subfigure}{0.49\columnwidth}
		\centering
		\includegraphics[width=\textwidth, angle=-90,trim=210 155 0 0,clip]{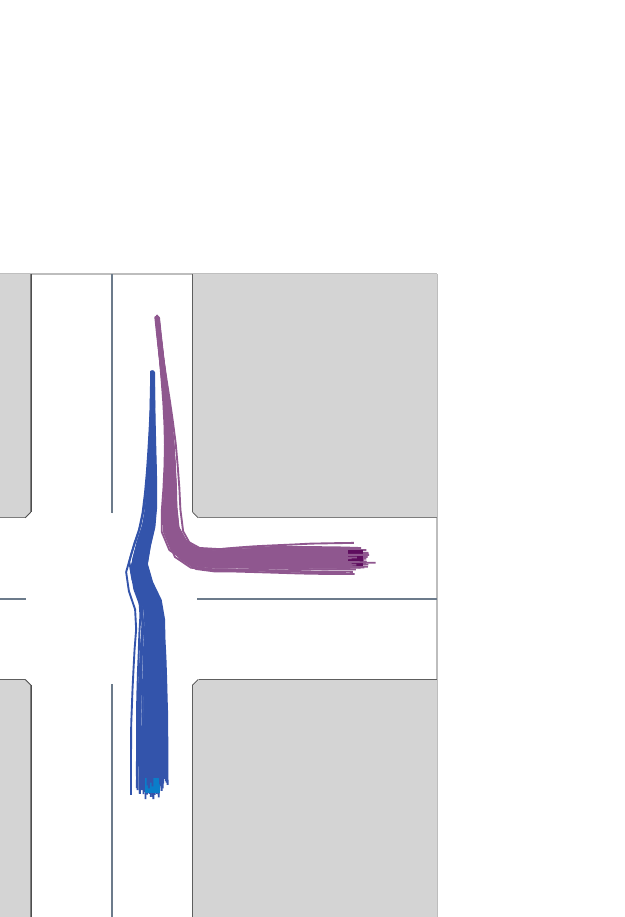}
		\vspace{-3.7em}
		\caption{Proposed} 
		\vspace{-0.5em}
	\end{subfigure}
	\caption{Visualization of 100 simulated trajectories of vehicle center in open-loop setting. (a) STRSv2. (b) CFSDMPC. (c) Proposed. The output trajectories of proposed method significantly reduces the uncertainty of the future trajectory.}
	\label{open_trajectory}
\end{figure}

\begin{figure}[htbp]
	\centering
	\begin{subfigure}{0.49\columnwidth}
		\centering
		\includegraphics[width=\textwidth,trim=108 55 95 45,clip]{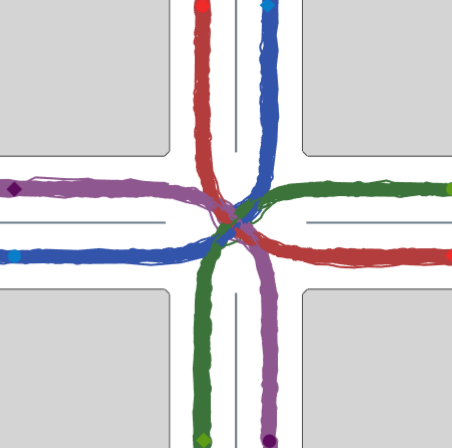}
		\vspace{-1.2em}
		\caption{STRSv2} 
            \vspace{0.5em}
	\end{subfigure}
	\centering
	\begin{subfigure}{0.49\columnwidth}
		\centering
		\includegraphics[width=\textwidth,trim=108 55 95 45,clip]{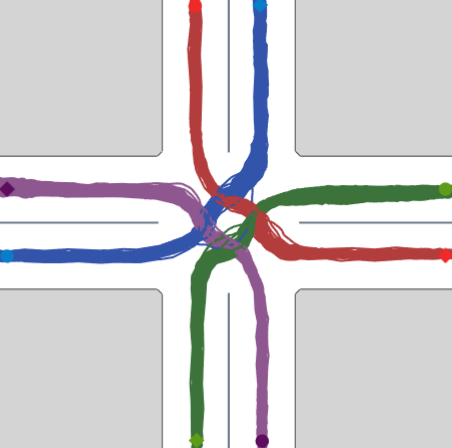}
		\vspace{-1.2em}
		\caption{CFSDMPC} 
            \vspace{0.5em}
	\end{subfigure}
	\centering
	\begin{subfigure}{0.49\columnwidth}
		\centering
		\includegraphics[width=\textwidth,trim=108 55 95 45,clip]{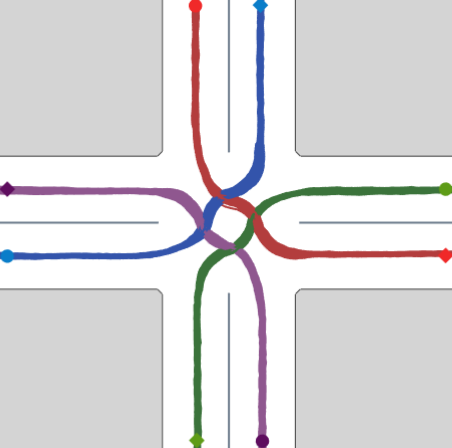}
		\vspace{-1.2em}
		\caption{Proposed} 
            \vspace{0.5em}
	\end{subfigure}
	\caption{Visualization of 100 simulated trajectories of vehicle center in closed-loop setting. (a) STRSv2. (b) CFSDMPC. (c) Proposed. The output trajectories of proposed method exhibits the smallest spread, which can reduce the potential collisions in multi-vehicle interactions.}
	\label{trajectory}
\end{figure}

\vspace{2em}
1) \emph{Trajectory Uncertainty and Safety Analysis:}

we first evaluate each method in an open-loop planning setting by performing 100 Monte Carlo simulations of intersection crossing. In this setting, two vehicles enter the intersection with initial state distributions of the same level, given by $\mathrm{diag}(0.4,0.4,0,0.2)$, $\mathrm{diag}(0.4,0.4,\pi/90,0.2)$, respectively. The resulting planned control sequence is executed forward without replanning. Fig. \ref{open_trajectory} presents the output trajectories of the three methods. It is evident that our approach achieves a significant reduction in the uncertainty of the future trajectory, thereby enhancing interaction safety. In contrast, the STRSv2 and CFSDMPC methods plan only nominal trajectories without directly controlling trajectory uncertainty, resulting in consistently large uncertainty throughout the entire interaction process.

Open-loop planning provides a transparent view of uncertainty evolution, while closed-loop replanning reflects operational behavior during real-time execution.
Thus, we further evaluate a closed-loop replanning setting in a challenging unprotected left-turn scenario, where four left-turning vehicles enter the intersection. We test the robustness of the proposed method through 100 Monte Carlo simulations of intersection crossing. Fig.~\ref{trajectory} illustrates the output trajectories of the three methods, while Table~\ref{tab2} records the number of collisions and the average speeds. It is obvious that our proposed method maintains smoother and more consistent trajectories than the STRSv2 and CFSDMPC methods while showing the smallest trajectory spread. This could lead to fewer potential collisions. According to Table~\ref{tab2}, the proposed method has the fewest collisions, recording 3 collisions, compared to 8 and 12 for STRSv2 and CFSDMPC methods, respectively. Additionally, our method achieves a higher average speed than the other two approaches. Our method demonstrates greater robustness to the motion uncertainties and sensory measurement noise. This robustness is attributed to our method's explicit consideration and active steering of the trajectory distributions during vehicle coordination, thereby providing a high level of safety regarding the uncertainty of the planned trajectory.

\begin{table}[h!]
	\captionsetup{font={small}}
	\caption{\centering{\scshape Comparison of Coordination Safety and Efficiency over 100 Monte Carlo Simulations}}
	\label{tab2}
	\centering
	\begin{tabular}{cccc}
		\toprule
		Metric & STRSv2\cite{li2024tightly} & CFSDMPC\cite{zhou2021distributed} & Proposed\\ \midrule
		Collisions (/100) & 8 & 12 & \textbf{3}\\
		Avg Speed (m/s) & 8.33  & 8.81 & \textbf{9.37}\\
        \bottomrule
	\end{tabular}
\end{table}

2) \emph{Speed Profile Analysis:}

\begin{figure}[htbp]
	\centering
	\begin{subfigure}{0.505\columnwidth}
		\centering
		\includegraphics[width=\textwidth,trim=16 3 33.5 19,clip]{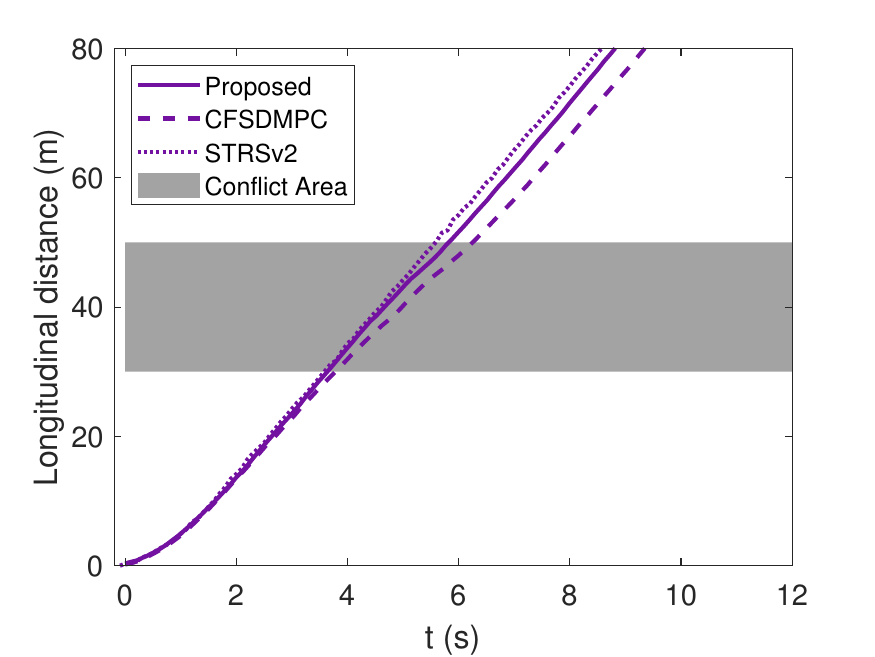}
		\vspace{-1.5em}
		\caption{Vehicle 1}	 \label{v1}
            \vspace{0.5em}
	\end{subfigure}
	\centering
	\begin{subfigure}{0.48\columnwidth}
		\centering
		\includegraphics[width=\textwidth,trim=34.5 3 33 19,clip]{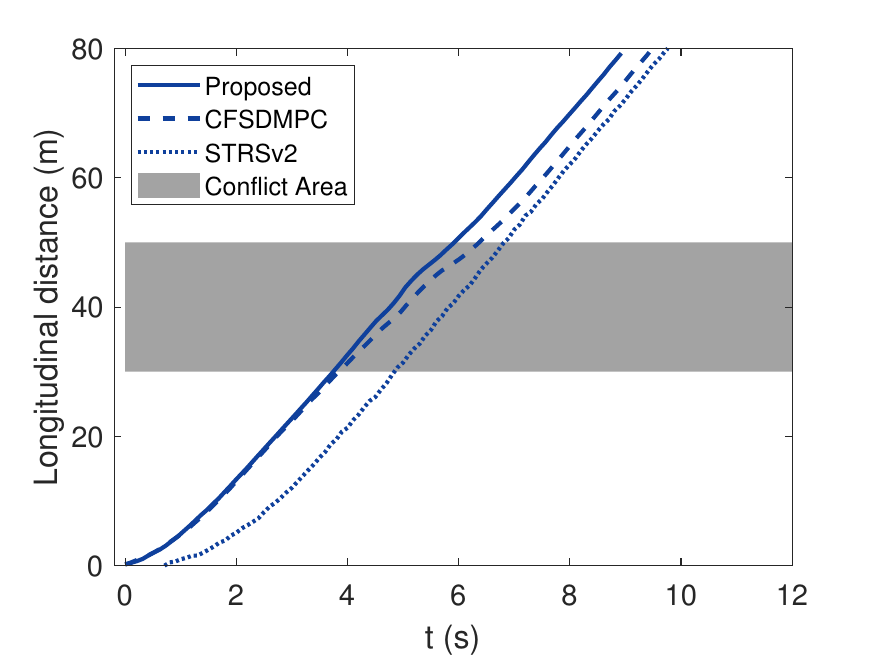}
		\vspace{-1.5em}
		\caption{Vehicle 2}	\label{v2}
            \vspace{0.5em}
	\end{subfigure}
	\centering
	\begin{subfigure}{0.505\columnwidth}
		\centering
		\includegraphics[width=\textwidth,trim=16 3 33.5 19,clip]{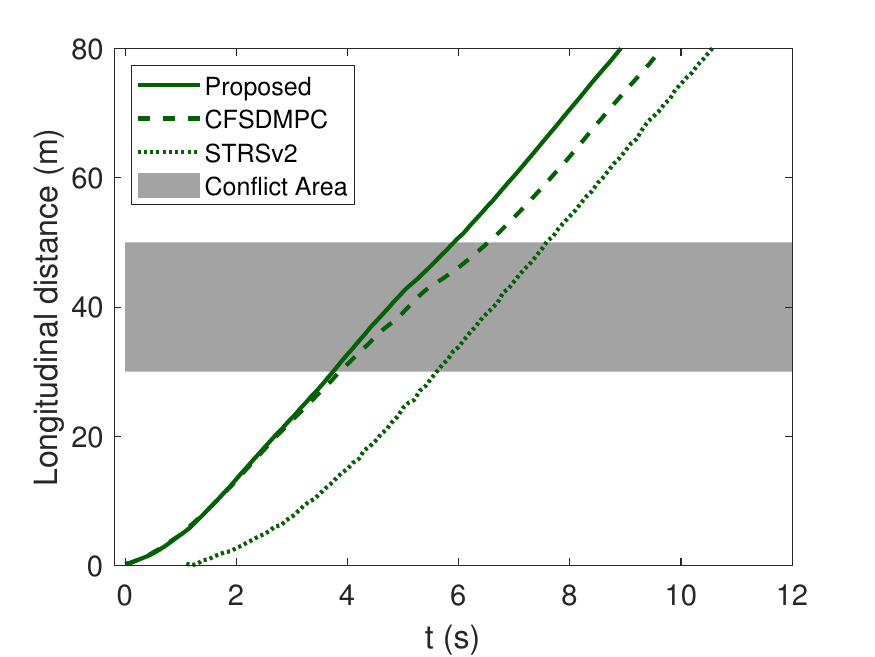}
		\vspace{-1.5em}
		\caption{Vehicle 3} \label{v3}
	\end{subfigure}
	\centering
    \begin{subfigure}{0.48\columnwidth}
     	\centering
    	\includegraphics[width=\textwidth,trim=34.5 3 33 19,clip]{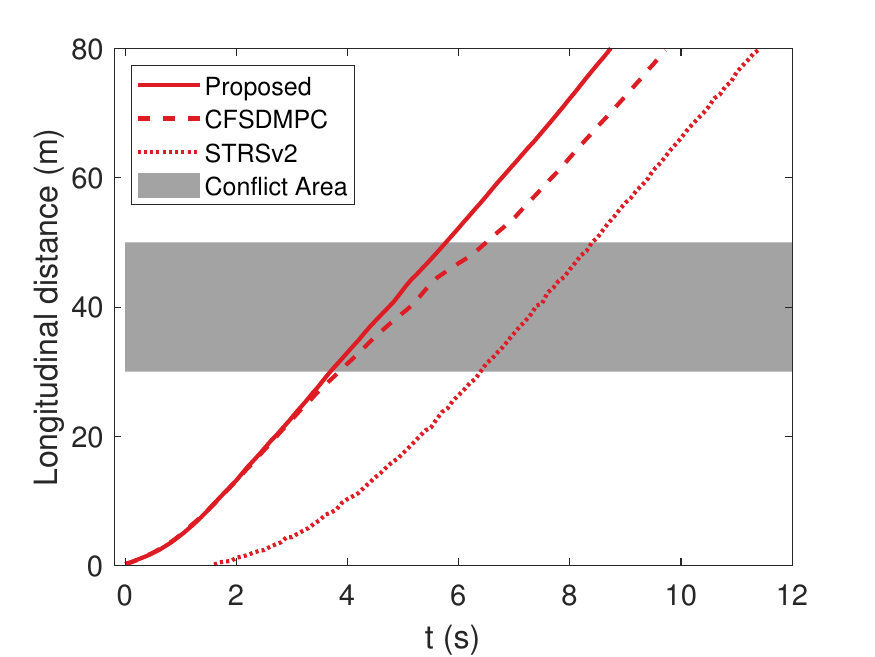}
    	\vspace{-1.5em}
    	\caption{Vehicle 4} \label{v4}
    \end{subfigure}
	\caption{Comparison of different methods for speed profile in ST coordinate system. (a) - (d) represent results for vehicle 1, vehicle 2, vehicle 3 and vehicle 4, respectively.}
	\label{speedprofile}
\end{figure}

Next, Fig.~\ref{speedprofile} presents the ST speed profile of the three methods in the Frenet coordinate system for one test case. The gray rectangular represents the conflict area in the center of the intersection. To facilitate a clear comparison, the four subfigures display the speed profile for the corresponding four vehicles. 
For vehicle 1, Fig.~\ref{speedprofile}(\subref{v1}) shows that the speed profiles of the three methods exhibit minor differences. The STRSv2 method demonstrates a slightly higher speed than the CFSDMPC and our method, as the sequential allocation of resource blocks in STRSv2 allows vehicle 1 to travel at maximum speed without accounting for other vehicles. 
For vehicles 2, 3, and 4, as shown in Fig.~\ref{speedprofile}(\subref{v2})-(\subref{v4}), the speed profiles of our approach show clear advantages over the CFSDMPC and STRSv2 methods, especially in the case of vehicle 4. 
In the STRSv2 method, vehicle passage times gradually increase, and in the CFSDMPC method, vehicle speeds visibly decrease when entering the conflict area. In contrast, the speed profiles of the four vehicles in our method rise more rapidly and show minimal differences, indicating that all four vehicles found an efficient passage solution.

\begin{figure}
	\centering
	\includegraphics[width=1.0\linewidth]{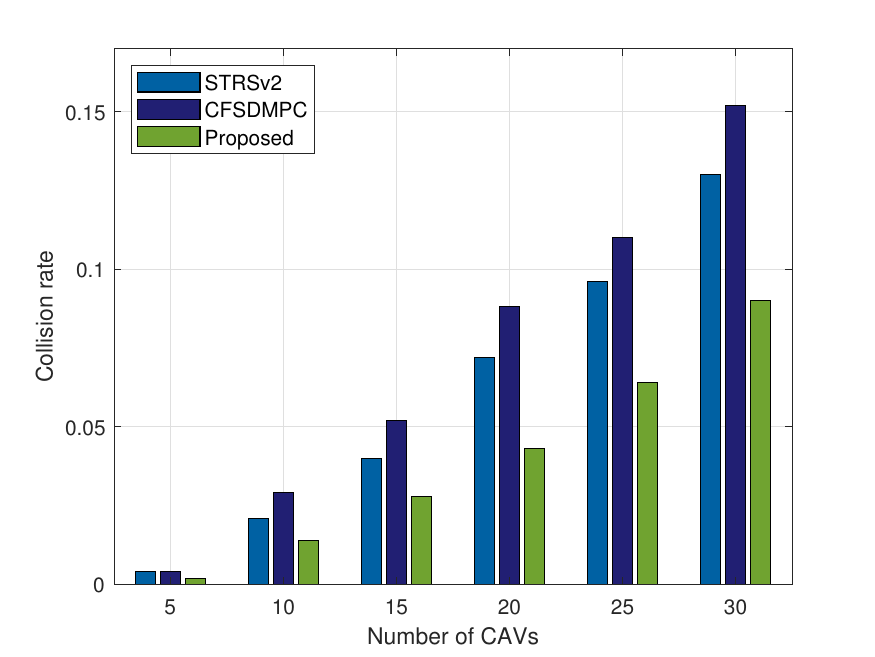}\\
	\vspace{-0.5em}
	\caption{Collision rate of different benchmarks.}\label{cr}
	\vspace{-1em}
\end{figure}

3) \emph{Collision Rate:}
We evaluated the crossing safety of different methods by scaling up the scenario with more vehicles. The actual performance of the compared methods cannot be fully demonstrated under a fixed vehicle flow. Therefore, we introduced two kinds of vehicle flow to reveal their intersection-crossing capabilities under varying vehicle patterns.
Following the vehicle patterns established in \cite{li2024tightly}, vehicle flow 1 consists of 50\% straight-going vehicles, 30\% left-turning vehicles, and 20\% right-turning vehicles. Vehicle flow 2 consists of 40\% left-turning vehicles, with straight-going and right-turning vehicles each accounting for 30\%.
Since our scenarios involve different vehicle patterns as well as random noise and disturbances, we conducted 200 Monte Carlo simulations for each scenario to mitigate potential biases and obtain collision rate statistics. Collision detection is performed using the Separating Axis Theorem (SAT) \cite{ericson2004real}, which efficiently checks the inter-vehicle overlap.

The collision rate of the three methods is presented in Fig.~\ref{cr}. The three methods achieve similar collision rates with a small number of vehicles. However, our method exhibits increasingly pronounced safety advantages as the number of vehicles increases. In the scenario with 30 vehicles, it reduces collision rates by 29.86\% and 40.79\% compared to the STRSv2 and CFSDMPC methods, respectively.
CFSDMPC method has the highest collision rate, as it does not explicitly account for vehicle dynamics and related uncertainties, making the execution trajectories prone to deviation. 
STRSv2 method avoids conflicts by preventing overlaps of space-time resource blocks, however, its safety margins fail to precisely accommodate error accumulation, leading to a relatively high collision risk.
The safety advantages of our method stem from precise characterization of uncertainty evolution throughout the entire coordination process. This enables the planning of the nominal trajectory while directly controlling the evolution of the trajectory distribution, thereby ensuring robust interactions among stochastic state trajectories. Additionally, the dynamic safety distance is adjusted based on the surrounding vehicle density. In other words, CAVs comprehensively account for their uncertainties and the surrounding environment to develop time-varying collision avoidance abilities, thereby ensuring robust safety guarantees.

\begin{figure}
	\centering
	\includegraphics[width=1.0\linewidth]{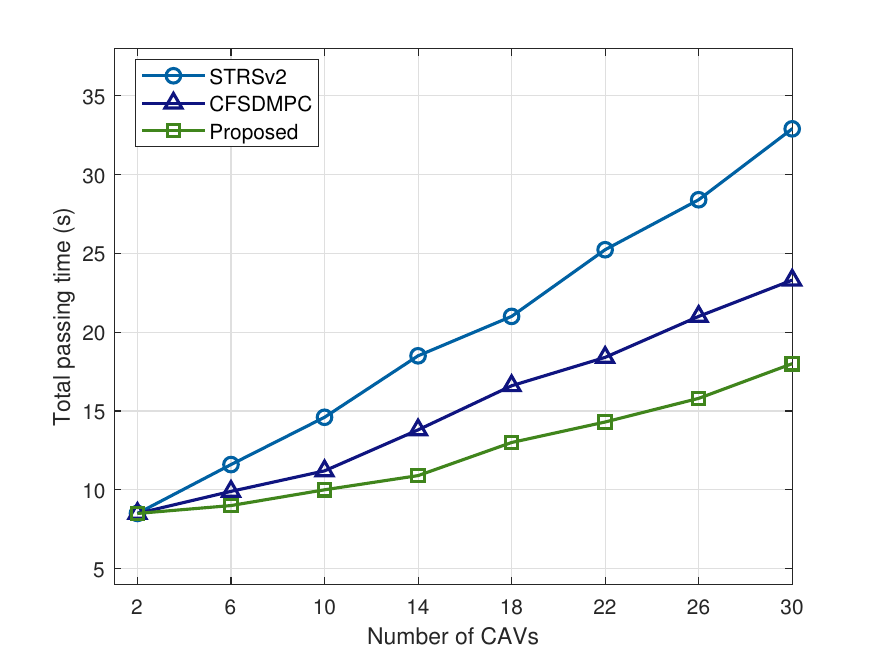}\\
	\vspace{-0.5em}
	\caption{Total passing time of different benchmarks.}\label{tpt}
	\vspace{-1em}
\end{figure}

4) \emph{Total Passing Time:}
Based on the same vehicle flows, Fig.~\ref{tpt} shows the total crossing time of the three methods with respect to the number of vehicles. 
It is evident that the total passing time of the STRSv2 method is significantly higher than that of CFSDMPC and our method. This is because STRSv2 method conducts sequential optimization for vehicles to allocate non-overlapping resource blocks, which results in suboptimal trajectories and reduced driving efficiency. 
In both CFSDMPC and our method, there is no decision delay between vehicles. However, our method achieves a shorter passing time compared to the CFSDMPC method. The CFSDMPC method requires more time and distance to converge to the reference trajectory due to the tracking deviation, and its distributed strategy can be considered a single-round negotiation process. In contrast, our method introduces a negotiation mechanism, which can improve the optimality of the solution through multiple negotiations.

 \begin{figure}
 	\centering
 	\includegraphics[width=1.0\linewidth]{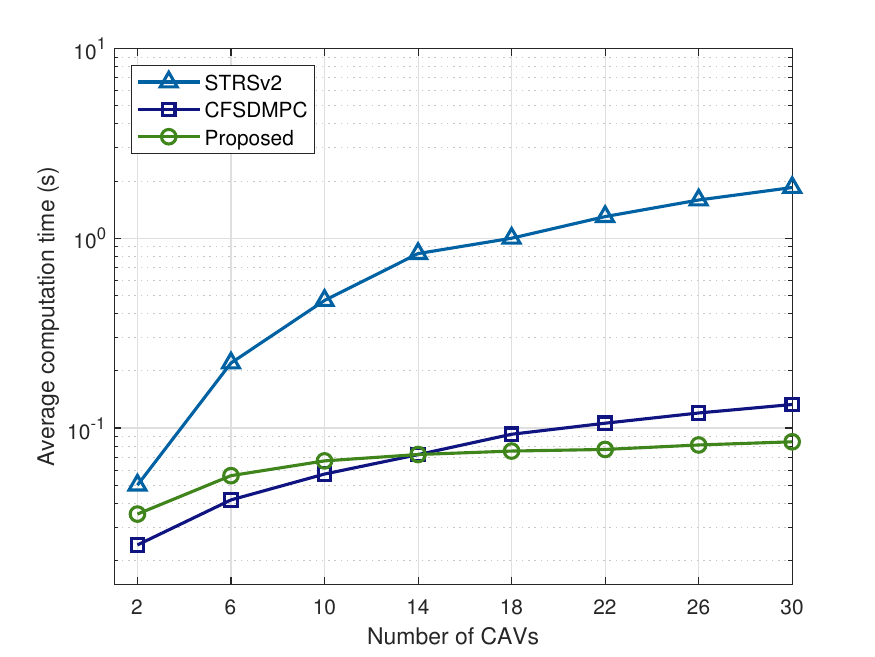}\\
 	\vspace{-0.5em}
 	\caption{Average computation time of different benchmarks.}\label{ct}
 	\vspace{-1em}
 \end{figure}
 
5) \emph{Computation Time:}
Fig.~\ref{ct} presents the computation time of the various methods under different numbers of vehicles. It is evident that our method and the CFSCMPC method have significantly shorter computation time compared to the STRSv2 method, due to time-consuming resource block overlap verification and allocation process in STRSv2. 
The CFSDMPC method, which approximates the problem to a sequence of quadratic programming, achieves a satisfactory computation time. The proposed method demonstrates comparable computation efficiency while exhibiting better scalability than the CFSDMPC method. Notably, when the number of vehicles exceeds 14, the computation time of our method is shorter than the CFSDMPC method.
This is because, as the number of vehicles increases, the CFSDMPC method requires more processing time to compute the convex feasible set in each iteration, leading to a faster increase in overall computation time. However, our non-convex collision constraint reformulation avoids iterative computations of feasible sets. Furthermore, our method leverages fully parallel computation and an efficient interactive attention mechanism, thus achieving superior computational efficiency and scalability.
    
\begin{table}[!t] 
	\captionsetup{font={small}}
	\caption{\centering{\scshape Ablation Study on Safety Formulation, ACT: Average Computation Time (s); TPT: Total Passing Time (s); CR: Collision Rate (\%)}}
	\label{tab3}
	\centering
	\begin{tabular}{cccc}
		\toprule
		  Setting & ACT (s)$\downarrow$ & TPT (s)$\downarrow$ & CR (\%)$\downarrow$  \\
		\midrule
		Uncertainty-aware Formulation & 0.0857 & \textbf{14.43} & \textbf{4.5}  \\
		Nominal-trajectory-only Formulation & \textbf{0.0346} & 16.85 & 8.0  \\
		\bottomrule
	\end{tabular}
\end{table} 
6) \emph{Ablation on the safety formulation:}
To evaluate the contribution of the proposed uncertainty-aware safety design, we compare the full uncertainty-aware formulation with a nominal-trajectory-only formulation while fixing the number of vehicles at 20. All other experimental and parameter settings remain unchanged. As shown in Table~\ref{tab3}, the uncertainty-aware formulation reduces the collision rate from 8.0\% to 4.5\%, corresponding to a relative reduction of 43.8\%, and also shortens the total passing time from 16.85~s to 14.43~s. 
This improvement stems from the more refined adaptive collision-avoidance constraints and the explicit control of trajectory distribution evolution during vehicle interactions, which allow the method to select relatively aggressive trajectories while still ensuring interaction safety in dense traffic.
Although the uncertainty-aware formulation incurs higher average computation time than the nominal-only variant, the additional cost remains moderate and is justified by the clear gains in safety and passing efficiency.

7) \emph{Ablation of parallel computation and interactive attention:}
The proposed parallel computation framework is compared with the corresponding centralized implementation. Table~\ref{tab4} presents the average computation time under different numbers of vehicles. Note that the collision rate and total passing time are not listed, as the three configurations yielded similar coordination performance. It can be seen that our method is an order of magnitude faster than the centralized approach and scales better with the number of vehicles. In our framework, convergence is generally achieved within 1-2 ADMM iterations, with single-iteration convergence occurring in approximately 80\% of cases.
Furthermore, the last two columns of Table~\ref{tab4} provide an ablation of the interactive attention mechanism through the computation-time comparison with and without attention. The results show that interactive attention reduces computational demand by around 7.4--15.4\%, with the benefit becoming more pronounced as the number of vehicles increases. In brief, by prioritizing the most critical neighboring vehicles and reducing coupling constraints, the proposed framework enables CAVs with interactive attention to solve local problems in parallel more efficiently, thereby further improving online computational performance.

\begin{table}[!t] 
	\captionsetup{font={small}}
	\caption{\centering{\scshape Ablation Study on Parallel Computation and Interactive Attention, Average Computation Time (ms); w/o and w/: Proposed Parallel Method without and with Interactive Attention}}
	\label{tab4}
	\centering
	\begin{tabular}{cccc}
		\toprule
		NUM CAV & Centralized &  Proposed(w/o)  & Proposed(w/) \\
		\midrule
		4  & 231.4   & 50.9   & \textbf{47.1}   \\
		8  & 871.7   & 71.7   & \textbf{63.9}  \\
		12 & 1648.5  & 83.6   & \textbf{72.6}   \\
		16 & 3768.2  & 93.3   & \textbf{80.2}   \\
        20 & 6573.9  & 101.4      & \textbf{85.7}   \\
		\bottomrule
	\end{tabular}
\end{table}

\begin{figure}[htbp]
	\centering
	\begin{subfigure}{0.49\columnwidth}
		\centering
		\includegraphics[width=\textwidth,trim=88 24 73 24,clip]{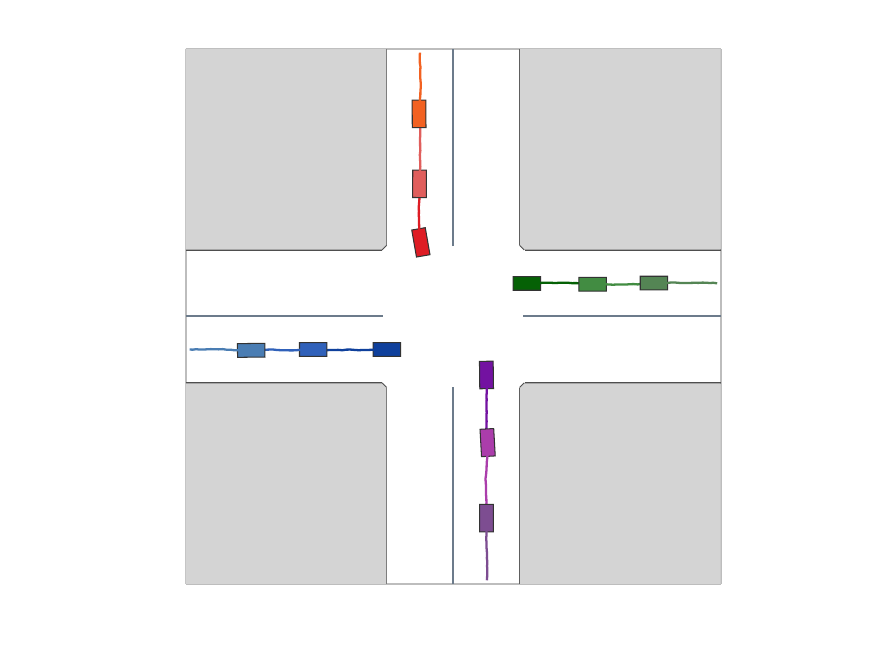}
		\vspace{-2em}
		\caption{$t=1.8$s}	\label{road1}
		\vspace{0.5em}
	\end{subfigure}
	\begin{subfigure}{0.49\columnwidth}
		\centering
		\includegraphics[width=\textwidth,trim=88 24 73 24,clip]{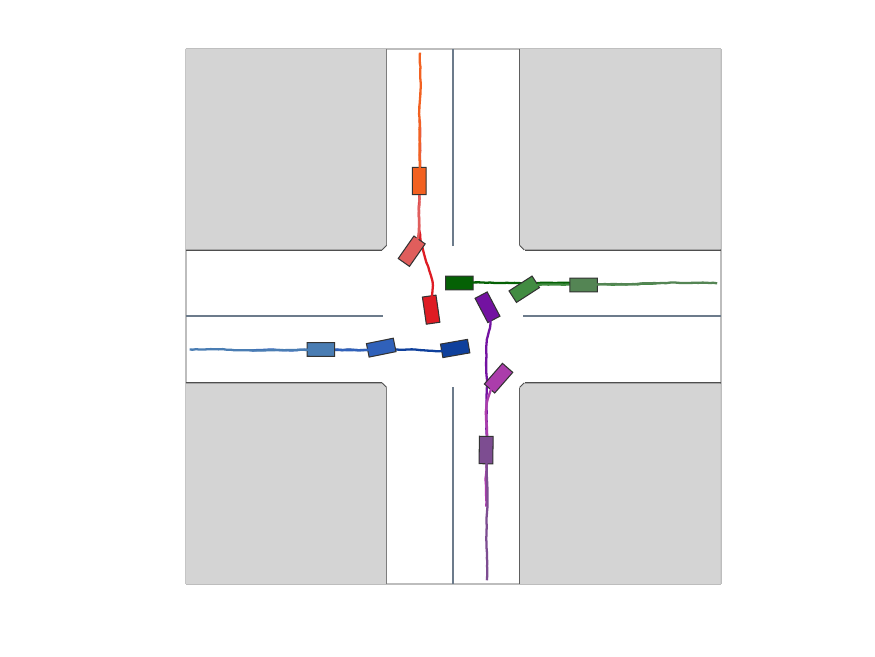}
		\vspace{-2em}
		\caption{$t=2.8$s} \label{road2}
		\vspace{0.5em}
	\end{subfigure}	
	\begin{subfigure}{0.49\columnwidth}
		\centering
		\includegraphics[width=\textwidth,trim=88 24 73 24,clip]{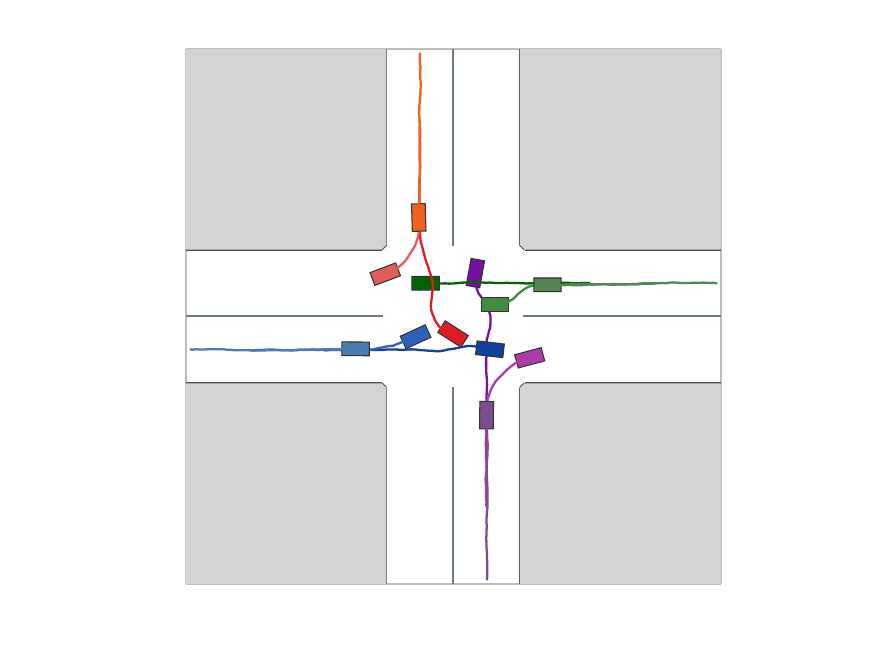}
		\vspace{-2em}
		\caption{$t=3.4$s}	\label{road3}
		\vspace{0.5em}
	\end{subfigure}
	\begin{subfigure}{0.49\columnwidth}
		\centering
		\includegraphics[width=\textwidth,trim=88 24 73 24,clip]{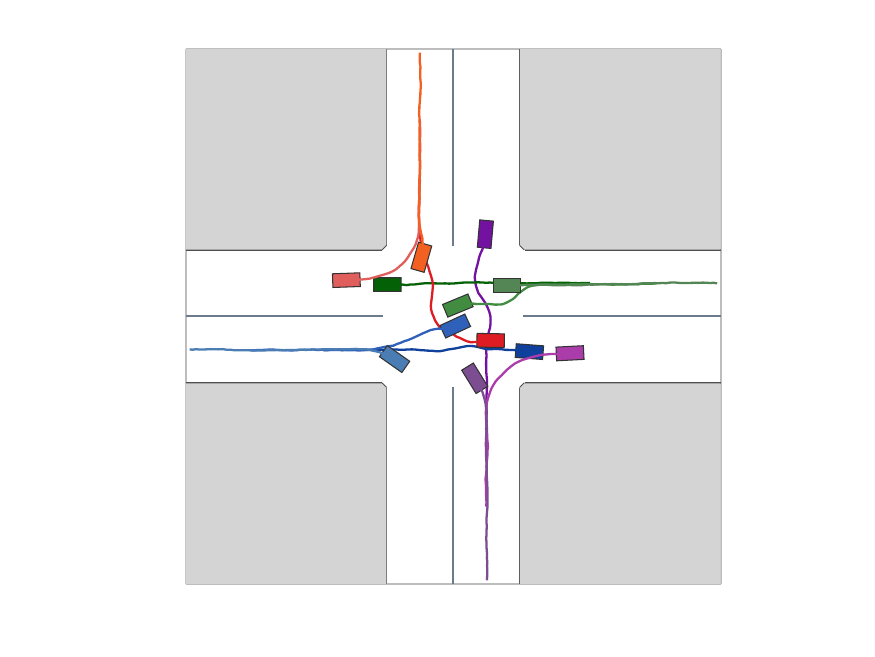}
		\vspace{-2em}
		\caption{$t=4.0$s} \label{road4}
		\vspace{0.5em}
	\end{subfigure}	
	\begin{subfigure}{0.49\columnwidth}
		\centering
		\includegraphics[width=\textwidth,trim=88 24 73 24,clip]{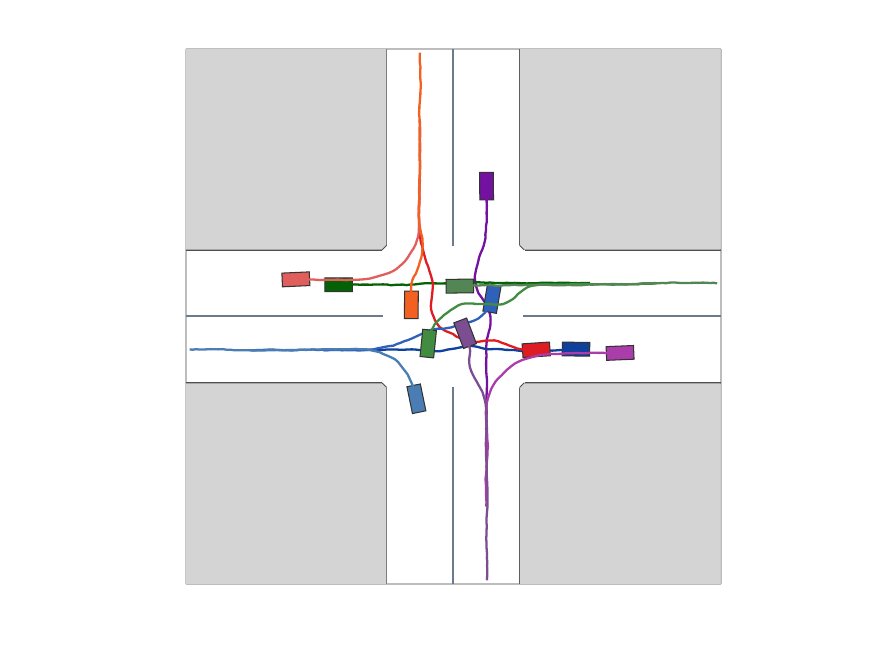}
		\vspace{-2em}
		\caption{$t=4.7$s}	\label{road5}
		\vspace{0.5em}
	\end{subfigure}
	\begin{subfigure}{0.49\columnwidth}
		\centering
		\includegraphics[width=\textwidth,trim=88 24 73 24,clip]{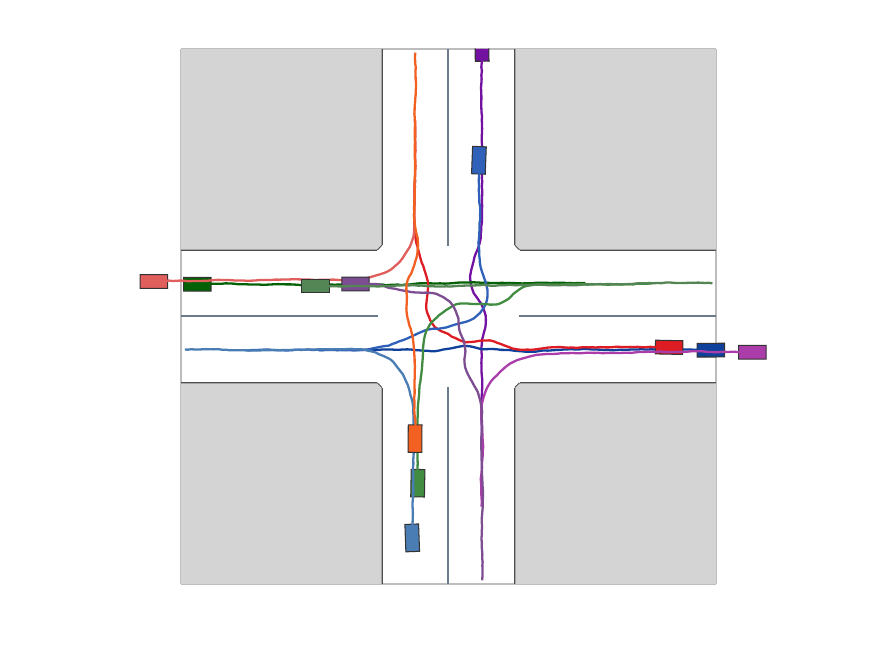}
		\vspace{-2em}
		\caption{$t=6.8$s} \label{road6}
		\vspace{0.5em}
	\end{subfigure}	
	\caption{All trajectories at different times for all vehicles with the proposed method.}
	\label{road}
\end{figure}

8) \emph{Qualitative Results:}
We demonstrate the efficacy of the proposed framework by illustrating the robust and safe passage at intersection. As shown in Fig.~\ref{road}, 12 vehicles simultaneously enter and navigate the intersection with different driving maneuvers. The solid black lines represent the road boundaries and centerlines to separate the lanes. 
Fig.~\ref{road}(\subref{road1})-(\subref{road6}) illustrates the driving process of all vehicles at different time stamps under motion and sensory measurement uncertainties. The solid lines of the same color behind each vehicle represent their history trajectories. 
Fig.~\ref{road} clearly shows that all vehicles successfully navigate the intersection with collision-free and smooth trajectories. The collision avoidance buffer dynamically adapts to the degree of state uncertainties and the surrounding vehicle density in the process. All vehicles interact flexibly to avoid each other, demonstrating robustness to disturbances and uncertainties.

\subsection{Discussion}\label{sectionVC}
1) \emph{Balancing safety assurance, real-time computation, and passing efficiency:}
    The proposed framework is fundamentally safety-oriented. It explicitly accounts for various sources of uncertainty and interference, and enforces an uncertainty-aware safety formulation throughout the coordination process. As a result, the evolution of trajectory distributions during multi-vehicle interactions can be directly regulated to enhance safety. Building on this safety-critical formulation, the framework achieves real-time performance through the fully parallel ADMM-DTN algorithm together with the interaction-selection mechanism, which jointly reduce computational burden and coupling scale without relaxing the safety requirements. Real-time replanning, in turn, improves practical safety by enabling timely responses to disturbances, uncertainty growth, and unexpected obstacles. At the same time, by precisely characterizing state uncertainty, the proposed method supports finer-grained safety interaction modeling with reduced conservatism, thereby improving passing efficiency when traffic conditions permit. Overall, the framework prioritizes safety-assured planning while simultaneously enabling efficient real-time computation and improved passing efficiency.

1) \emph{Handling Human-Driven Vehicles (HDVs):}
    The proposed framework primarily focuses on uncertainty-aware trajectory coordination and real-time parallel computation. Nevertheless, it can accommodate HDVs at a coarse level by modeling them as moving obstacles and enforcing the corresponding collision-avoidance constraints. Existing approaches for handling human drivers in coordinated autonomy mainly fall into two categories: intent inference and behavior prediction \cite{wang2024distributed}, and utility- or social-preference-based behavior modeling \cite{le2022cooperative}. Accordingly, a finer-grained treatment of HDVs could be enabled by incorporating a dedicated human-driver modeling module, such as intent inference, trajectory prediction, or preference-aware behavior modeling. Specifically, the resulting probabilistic HDV trajectory distributions can be integrated into the existing safety constraints, whereas inferred utilities or preferences can be used to parameterize game-theoretic interaction formulations.

\section{Real-world Experiments}\label{sectionVI}

\begin{figure}[htbp]
	\centering
	\begin{subfigure}{0.498\columnwidth}
		\centering
		\includegraphics[width=\textwidth]{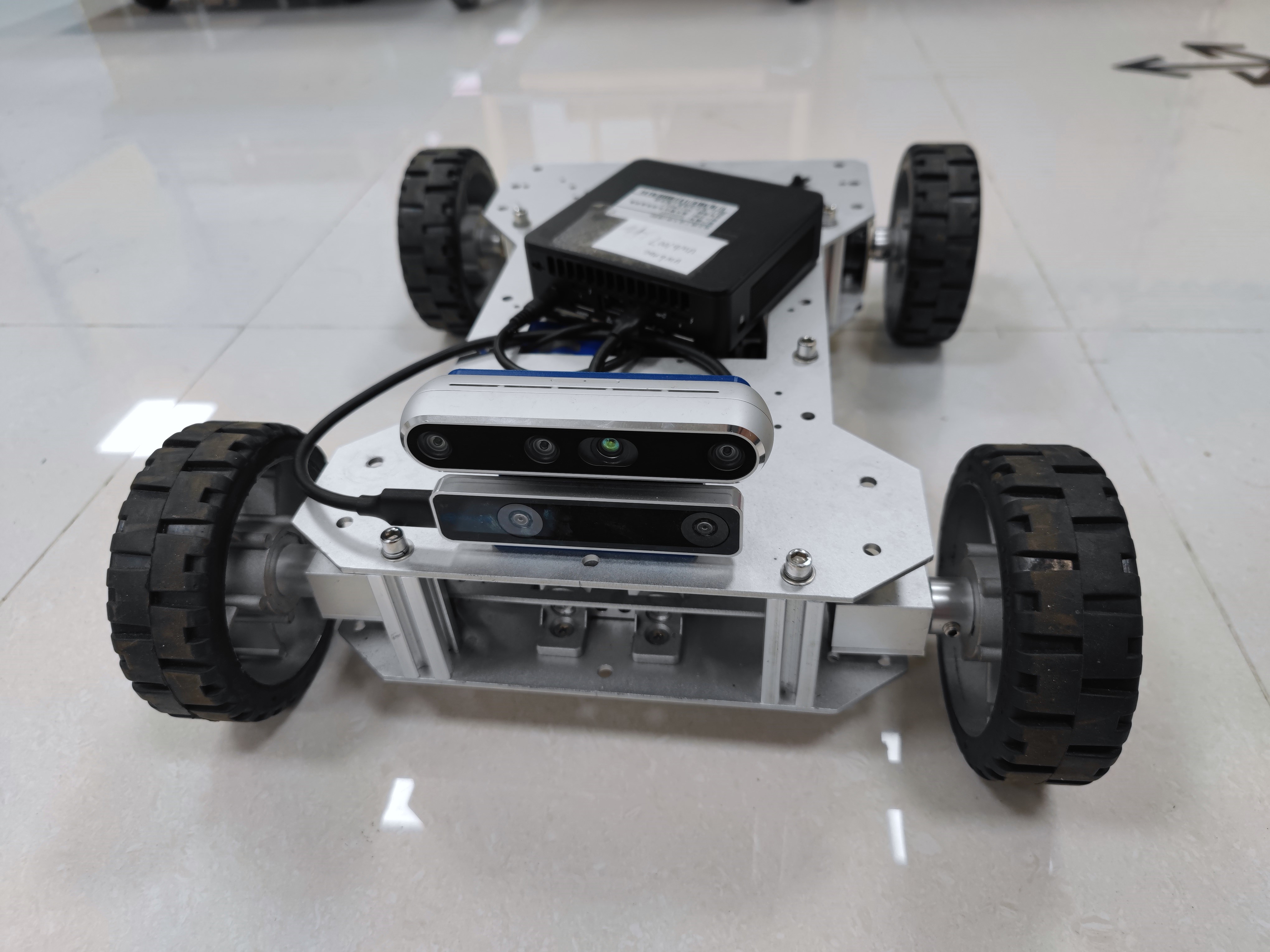}
	\end{subfigure}
	\begin{subfigure}{0.485\columnwidth}
		\centering
		\includegraphics[width=\textwidth]{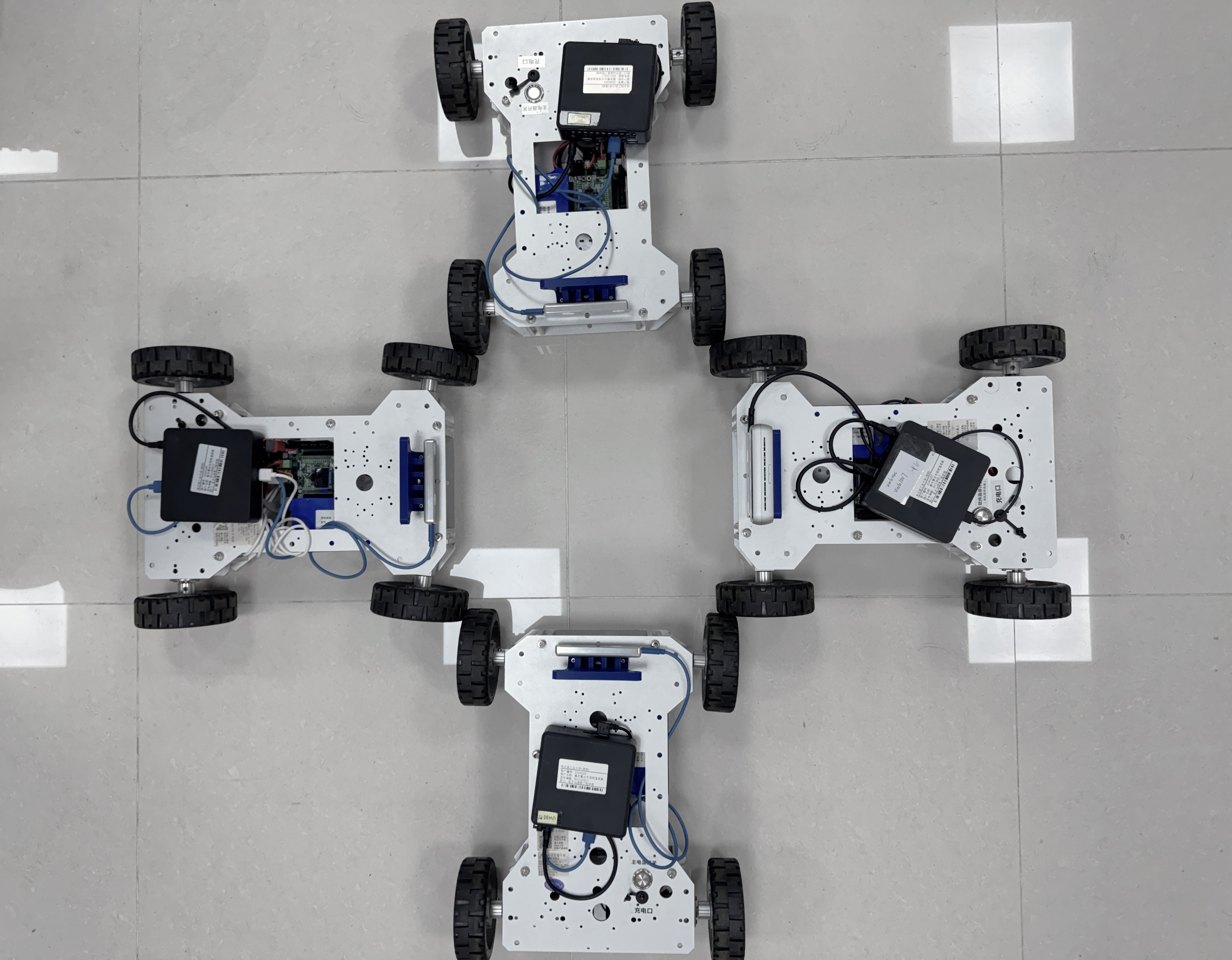}
	\end{subfigure}	
	\caption{Autonomous mobile robots used in the experiments.}\label{ackerman}
\end{figure}

Finally, to further validate the practicality and robustness of the proposed framework in a real-world environment, we implement the algorithm on four autonomous mobile robots with the Robot Operating System (ROS). 
In the experiment, the robots used are based on the platform from\cite{li2024tightly} as shown in Fig.~\ref{ackerman}, which is equipped with the Intel D455 depth camera and the T265 tracking camera to perceive the environment and measure the robot's position and velocity.     Each mobile robot uses the Intel NUC 12 Pro microcomputer with Intel i7-1260P 2.10 GHz CPU and 16 GB of RAM as the onboard computing platform to process sensor data and compute local trajectories. The robots share trajectory information with the neighboring robots via Wi-Fi links.
The disturbance matrix $G$ is estimated online using the deviation of the physical system from the nominal model. At each sampling time $t$, the disturbance matrix $G$ is updated by the square root of the sample covariance of the most recent deviation, similar to the approach in\cite{knaup2023safe}. The measurement noise matrix is set as $D = \mathrm{diag}(0.1,0.1,\pi/180,0.1)$ based on the precision evaluated in\cite{bayer2019autonomous}.
The vehicle parameters are as follows: $l_{\text {car}}=0.45$m, $w_{\text {car}}=0.38$m, $v_{\text {max}}=0.6$m/s, $a_{\text {max}}=0.1$m/s$^2$.

\begin{figure}
	\centering
	\includegraphics[width=1\linewidth]{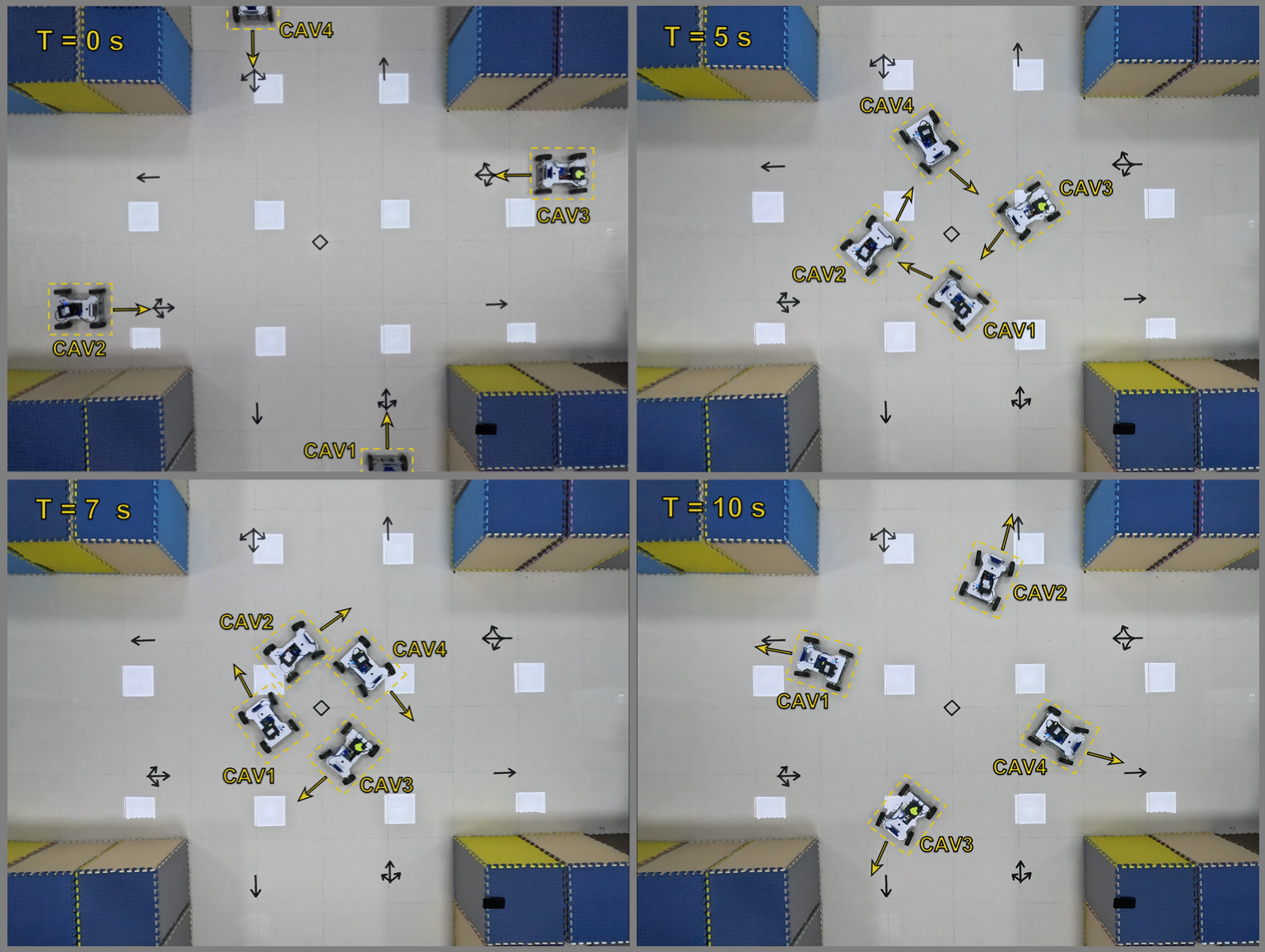}\\
	\caption{Unprotected left-turn intersection crossing experiments.}\label{crossingexp}
	\vspace{-1em}
\end{figure}
Fig.~\ref{crossingexp} shows a typical laboratory intersection scenario with sizes of 8.4 m × 8.4 m, and the lane width is 1.2m. It can be viewed as a size-scaled road intersection. In this experiment, four CAVs perform unprotected left-turn maneuvers. From the Fig.~\ref{crossingexp}, we can observe that all four CAVs can enter the intersection simultaneously, which greatly improves crossing efficiency. They can flexibly and robustly avoid collisions at each time step despite disturbances and noise. Finally, all four CAVs successfully and safely navigate the intersection along stable trajectories.

\begin{figure}
	\centering
	\includegraphics[width=1\linewidth]{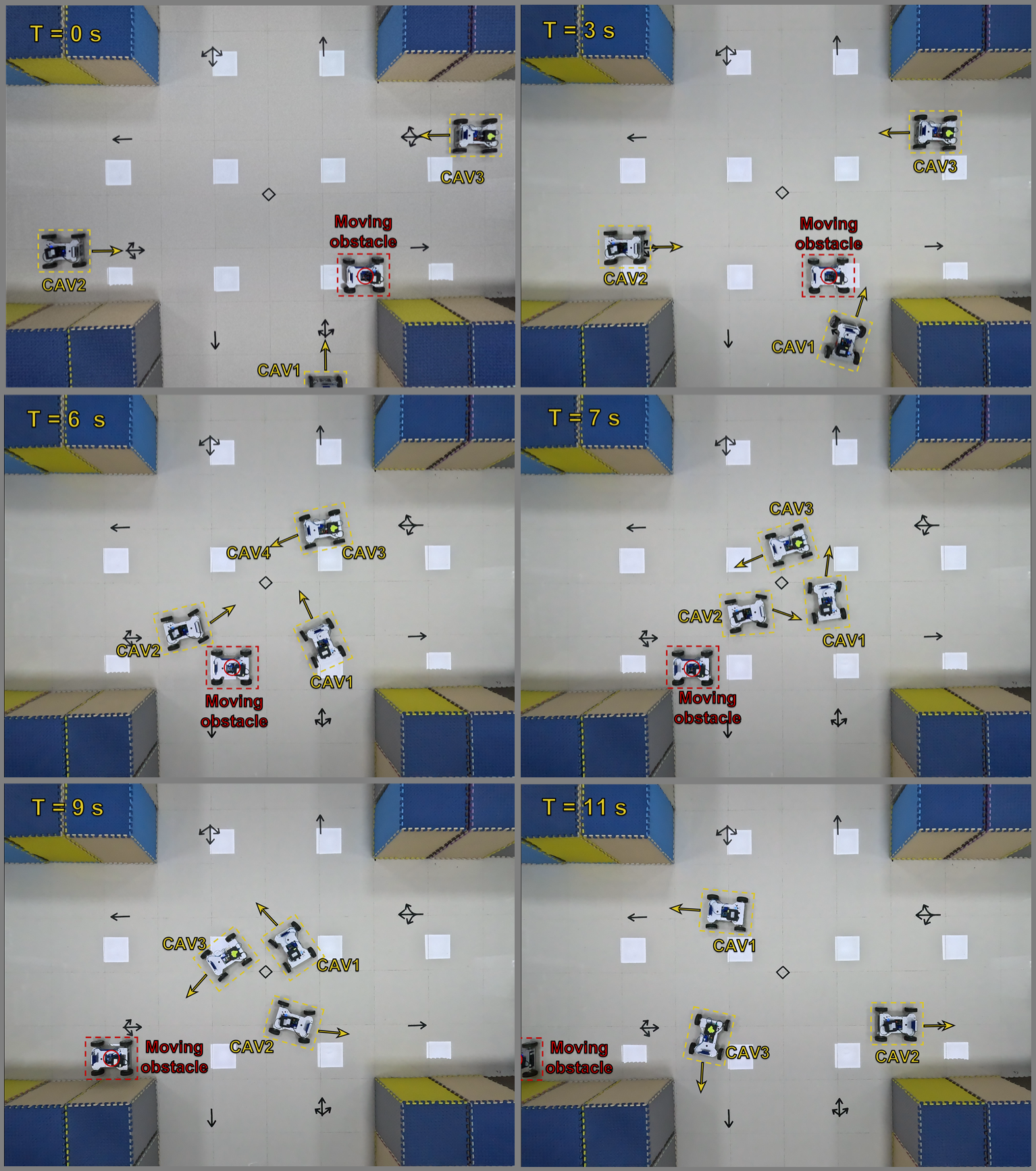}\\
	\caption{Intersection crossing experiments with unexpected dynamic obstacle.}\label{avoidanceexp}
	\vspace{-1em}
\end{figure}

To show the ability of the proposed framework to handle unexpected non-CAV vehicles during intersection crossing, we introduced an unexpected moving obstacle on the road, marked as a red box. This dynamic obstacle accelerates along the road within a speed range of 0.18 - 0.6 m/s and may appear in the CAVs' paths. It does not communicate with other CAVs and thus can be regarded as a human-driven or legacy vehicle. 
To avoid potential collisions, the three CAVs entering the intersection need to not only avoid each other but also evade the unexpected moving obstacle. Fig.~\ref{avoidanceexp} shows the experimental results. From $t=3$s to $t=9$s, it is clear that CAV1 and CAV2 can safely navigate around the unexpected obstacle upon spotting it. Meanwhile, the three CAVs safely interact to avoid each other during navigation.
This shows that the proposed framework is capable of simultaneously handling collision avoidance among CAVs and with unexpected dynamic obstacles, showcasing the advanced adaptability of cooperative driving.

\begin{table}[!t] 
	\captionsetup{font={small}}
	\caption{\centering{\scshape Statistics of the Real-World Experiments, ACT: Average Computation Time (ms); WCT: Worst-case Computation Time (ms); MSD: Minimum Separation Distance (cm); SR: Success Rate (\%)}}
	\label{tab5}
	\centering
	\begin{tabular}{ccccc}
		\toprule
		Experiment & ACT (ms) & WCT (ms)  & MSD (cm) & SR (\%)\\ 
        \midrule
		Unprotected Left-turn & 72.8 & 95.4  & 10.2 & 98.0 \\
		Unexpected Obstacle  & 76.6 & 104.9  & 8.8  & 96.0\\
        \bottomrule
	\end{tabular}
\end{table}
To complement the visualization results shown in Figs.~\ref{crossingexp} and \ref{avoidanceexp}, we further repeated each real-world experiment 50 times. Table~\ref{tab5} reports the average computation time (ACT), worst-case computation time (WCT), minimum separation distance (MSD), and success rate (SR) for the two scenarios.
For the unprotected left-turn experiment, these results indicate stable and safe intersection crossing while also confirming efficient online computation in the physical system.
Compared with the unprotected left-turn case, the unexpected-obstacle experiment exhibits a slightly lower SR and a smaller MSD, due to additional dynamic non-CAV vehicles and the resulting unanticipated interactions. Although the worst-case computation time slightly exceeds the 0.1 s replanning interval, this occurs rarely. In such cases, the system continues executing the previous feasible solution until the updated plan becomes available. Overall, the results still demonstrate robust operation, stable safety margins, and practical online performance in this more challenging mixed-traffic scenario.

Our framework is explicitly designed to adapt to practical deployments with varying computational resources: (i) the proposed ADMM-DTN is fully parallel across agents, and (ii) the Jacobi-style negotiation guarantees a feasible solution for any number of negotiation rounds, enabling the system to trade off solution quality versus available compute. (iii) the decision variables are organized sequentially rather than in a batch form, which substantially accelerates computation \cite{pilipovsky2024computationally}.

\section{Conclusion}\label{sectionVII}

This paper proposes a robust and real-time coordination framework for cooperative trajectory planning of CAVs in dynamic and uncertain environments. A novel uncertainty-aware coordination problem formulation is developed to directly control trajectory distribution evolution during vehicle interactions, and is solved using a fully parallel ADMM-based distributed trajectory negotiation algorithm (ADMM-DTN) and an interactive attention mechanism.
The coordination framework adapts to various time-varying scenarios while simultaneously ensuring probabilistic safety guarantees and efficient real-time implementation.
Comprehensive simulations and experiments have shown that the proposed framework outperforms representative benchmarks in terms of safety and efficiency. The collision rates can be reduced by up to 29.86\% and 40.79\% compared to various benchmark methods. Furthermore, the computation time is significantly shorter by an order of magnitude compared to STRSv2 and exhibits better scalability than CFSDMPC.
Specifically, our framework can robustly and flexibly handle unexpected non-CAV obstacles in real-world experiments, addressing the applicability limitations of most existing vehicle coordination methods.

Future research directions include extending the proposed framework to explicitly model and predict human-driver social behaviors, as well as reason over such behaviors in a game-theoretic manner—moving beyond non-communicating obstacle handling—to further broaden its applicability.

\bibliographystyle{IEEEtran} 
\bibliography{Acceleratedreference}

\appendices
\section{Bicycle Model and Dynamics Linearization and Discretization} 
\label{bicycle}

	Based on the bicycle model, the deterministic vehicle dynamics of CAVs follows the nonlinear kinematic function
	\begin{equation}
	\left[\begin{array}{c}
	\dot{\mathrm{x}}_{i} \\
	\dot{\mathrm{y}}_{i} \\
	\dot{\theta}_{i} \\
	\dot{\mathrm{v}}_{i}
	\end{array}\right]=\left[\begin{array}{c}
	v_{i} \cos \left(\theta_{i}\right) \\
	v_{i} \sin \left(\theta_{i}\right) \\
	\frac{v_{i}}{L_w} \tan \left(\delta_{i}\right) \\
	a_{i}
	\end{array}\right],
	\end{equation}
	where $(\dot{\bullet}):=\frac{\partial}{\partial t}(\bullet)$ and $L_w$ denotes the distance between the front and rear wheels axles.
	Using the first-order Taylor polynomial, this nonlinear function can be linearized around a nominal trajectory $\bar{x}_i=\left[\bar{x}_i^0, \ldots, \bar{x}_i^N\right]^{\top}$ and $\bar{u}_i=\left[\bar{u}_i^0, \ldots, \bar{u}_i^{N-1}\right]^{\top}$, resulting in
	\begin{equation}
	\dot{x}_{i}=A_{i} x_{i} + B_{i} u_{i} +r_{i},
	\end{equation}
	where the associated coefficients $(A_{i},B_{i},r_{i})$ are derived as

	\begin{equation}\label{linear}
	A_{i}=\!\!\left[\begin{array}{cccc}
	0 & 0 &\!\!\!\!  \!\cos\! \left(\bar{\theta}_{i}\right) &\!\!\!\!\!\!\!\!\!\! - \bar{v}_{i,t}\! \sin \!\left(\bar{\theta}_{i}\right) \\
	0 & 0 &\!\!\!\! \! \sin\! \left(\bar{\theta}_{i}\right) &\!\!\!\!\!\!\!\!\!\!  \bar{v}_{i,t}\! \cos\! \left(\bar{\theta}_{i}\right) \\
	0 & 0 &  \frac{1}{L_w} \tan \left(\bar{\delta}_{i}\right) &\!\!\!\!\!\!\!\!\!\! 0 \\
	0 & 0 &\!\!\! 0 &\!\!\!\!\!\!\!\!\!\! 0 
	\end{array}\!\!\!\!\right], 
	\end{equation}
	\begin{equation}
	B_{i}=\left[\begin{array}{cc}
	0 &\! 0 \\
	0 &\! 0 \\
	0 &\! \frac{ \bar{v}_{i}}{L_w \cos ^2\left(\bar{\delta}_{i}\right)} \\
	1 &\! 0 
	\end{array}\right],	\quad 
	r_{i}=\left[\begin{array}{cc}
	\bar{\theta}_{i} \bar{v}_{i} \sin(\bar{\theta}_{i})  \\
	-\bar{\theta}_{i} \bar{v}_{i} \cos(\bar{\theta}_{i})  \\
	\frac{ \bar{\delta}_{i} \bar{v}_{i}}{L_w \cos ^2\left(\bar{\delta}_{i}\right)} \\
	0  
	\end{array}\right].
	\end{equation}
	
	Subsequently, to ensure high accuracy and numerical stability, 2nd-order Runge-Kutta integration is employed to discretize the dynamics. The discrete-time interval is denoted as $\tau$. The stochastic and linear time-varying dynamics with noise added is constructed as follows
    \begin{equation}
    x_i^{k+1}=A_i^k x_i^k+B_i^k u_i^k+r_i^k +G_i^k w_i^k,
    \end{equation}
	where 
	\begin{equation}
	A_i^k=I+\tau  A_{i}+\frac{\tau^2}{2} A_{i}^2,
	\end{equation}
	\begin{equation}
	B_i^k=\tau B_{i}+\frac{\tau^2}{2} A_{i}B_{i},
	\end{equation}
	\begin{equation}
	r_i^k=\tau r_{i}+\frac{\tau^2}{2} A_{i}r_{i},
	\end{equation}
    and the diagonal elements of $G_i^k$ correspond to the magnitude of the noise.

\section{Proof of Lemma 1}
\label{proof}
Proof. It follows from that the closed-loop covariance propagation is $\kappa$-contractive in the positive semidefinite order through the computed feedback gain
    \begin{equation}\label{50}
    \left(A_i^k+B_i^k K_i^k\right) \hat{P}_i^k\left(A_i^k+B_i^k K_i^k\right)^{\top} \preceq \kappa^2 \hat{P}_i^k, 
   \end{equation}
   Let $W_i^k$ denote the filtered noise term in (\ref{sigmadynamic}), i.e., $W_i^k:=L_i^{k+1} P_{\xi, i}^{k+1} L_i^{k+1 \top} \succeq 0$ and let $P_{\infty}$ denote the steady-state covariance achieved when the system converges.
   Substituting (\ref{50}) into (\ref{sigmadynamic}), the evolution of the covariance is bounded as:
   \begin{equation}\label{51}
   \hat{P}_i^{k+1} \preceq \kappa^2 \hat{P}_i^k+W_i^k . 
   \end{equation}
   We now iterate the covariance evolution over time:
   \begin{equation}
   \hat{P}_i^{k} \preceq \kappa^{2 k} \hat{P}_i^0+\sum_{t=0}^{k-1} \kappa^{2 (k-t-1)} W_i^t, 
   \end{equation}
   the noise term $W_i^k$ is a bounded filtered noise term, satisfying $W_i^k \preceq \bar{W}$, we have, 
   \begin{equation}
    \hat{P}_i^k \preceq \kappa^{2 k} \hat{P}_i^0+\frac{1}{1-\kappa^2} \bar{W} , 
    \end{equation}
    therefore, as $k \rightarrow \infty$, the covariance converges to the steady-state value:
    \begin{equation}
    \hat{P}_i^{\infty} \preceq \frac{1}{1-\kappa^2} \bar{W}=P_{\infty} . 
    \end{equation}
    From (\ref{51}), we have
    \begin{equation}
    \hat{P}_i^{k+1}-\hat{P}_i^k \preceq\left(\kappa^2-1\right) \hat{P}_i^k+\bar{W} , 
    \end{equation}
    since $\hat{P}_i^k \succeq P_{\infty}=\frac{1}{1-\kappa^2} \bar{W}$, then $\bar{W} \preceq\left(1-\kappa^2\right) \hat{P}_i^k$. Substituting,
    \begin{equation}
    \hat{P}_i^{k+1}-\hat{P}_i^k \preceq\left(\kappa^2-1\right) \hat{P}_i^k+\left(1-\kappa^2\right) \hat{P}_i^k =0. 
    \end{equation}
    Therefore, if $\hat{P}_i^0 \succeq P_{\infty}$, then $\hat{P}_i^{k+1}\preceq\hat{P}_i^k$ for all $k$, and the covariance will monotonically decrease toward the steady-state bound $P_{\infty}$.

\end{document}